\pdfoutput=1
\documentclass[11pt]{article}

\usepackage[final]{acl}

\usepackage{times}
\usepackage{latexsym}

\usepackage[T1]{fontenc}
\usepackage[utf8]{inputenc}

\usepackage{microtype}
\usepackage{comment}
\usepackage{inconsolata}

\usepackage{graphicx}
\usepackage{array}
\usepackage{multirow}
\usepackage{booktabs}
\usepackage{multirow}
\usepackage{graphicx}
\usepackage{subcaption}
\usepackage{amsmath}
\usepackage{booktabs}
\usepackage{listings}
\usepackage{xcolor}
\usepackage{tabularx}
\usepackage{booktabs}
\usepackage{colortbl}
\usepackage{soul}
\usepackage{pifont}

\title{CAMI: A Counselor Agent Supporting Motivational Interviewing through State Inference and Topic Exploration}

\author{
 \textbf{Yizhe Yang\textsuperscript{1}~\thanks{Work was done during a visit at SMU.}},
 \textbf{Palakorn Achananuparp\textsuperscript{2}},
 \textbf{Heyan Huang\textsuperscript{1}~\thanks{Corresponding Author}},
 \textbf{Jing Jiang\textsuperscript{3}},
\\
 \textbf{Kit Phey Leng \textsuperscript{4}},
 \textbf{Nicholas Gabriel Lim \textsuperscript{5}},
 \textbf{Cameron Tan Shi Ern \textsuperscript{6}},
 \textbf{Ee-peng Lim\textsuperscript{2}}
\\
\\
 \textsuperscript{1}Beijing Institute of Technology,
 \textsuperscript{2}Singapore Management University,
 \textsuperscript{3}Australian National University, 
 \\
 \textsuperscript{4}National Institute of Education,
 \textsuperscript{5}Singapore University of Social Sciences,
 \textsuperscript{6}National University of Singapore
\\
 \small{
   \{yizheyang,hhy63\}@bit.edu.cn, \{palakorna,eplim\}@smu.edu.sg, jing.jiang@anu.edu.au
 }
}

\begin{document}
\maketitle
\begin{abstract}
Conversational counselor agents have become essential tools for addressing the rising demand for scalable and accessible mental health support. This paper introduces 
CAMI, a novel automated counselor agent grounded in Motivational Interviewing (MI) -- a client-centered counseling approach designed to address ambivalence and facilitate behavior change.  
CAMI employs a novel STAR framework, consisting of client's state inference, motivation topic exploration, and response generation modules, leveraging large language models (LLMs). These components work together to evoke change talk, aligning with MI principles and improving counseling outcomes for clients from diverse backgrounds.
We evaluate CAMI’s performance through both automated and manual evaluations, utilizing simulated clients to assess MI skill competency, client's state inference accuracy, topic exploration proficiency, and overall counseling success. 
Results show that CAMI not only outperforms several state-of-the-art methods but also shows more realistic counselor-like behavior. Additionally, our ablation study underscores the critical roles of state inference and topic exploration in achieving this performance.

\end{abstract}

\section{Introduction}

{\bf Motivation.} Motivational Interviewing (MI) is a client-centered counseling technique aimed at addressing ambivalence and facilitating behavior change in clients, particularly for issues like binge drinking and substance use~\citep{miller2002motivational}. 
%By addressing ambivalence and enhancing self-efficacy, 
MI promotes intrinsic motivation and fosters a collaborative relationship between clients and counselors, ultimately improving clients' commitment to intervention programs~\citep{martins2009review}. 
%The effectiveness of MI techniques relies on psychotherapy skills that include not only dialogue strategies but also personalized counseling programs and change talk exploration throughout various stages of the counseling process~\citep{cole2023using}.

%The recent emergence of large language models (LLMs)~\citep{brown2020language,ouyang2022training,achiam2023gpt,dubey2024llama} presents new opportunities for generating diverse, flexible, and engaging dialogues. 
Although recent works building LLM-based counselor agents have shown some promising results, we observe at least three limitations with existing works. First, they do not follow the collaborative counseling style in MI. For example, \citet{steenstra2024virtual} describes MI in a prompt which may not provide LLM adequate guidance to perform a complex multi-turn conversational task such as MI counseling. Second, those works that use strategy planning overlook the client's state and the need to explore a diverse range of topics that can evoke change talk based on the client’s unique motivations~\citep{hsu2023helping,sun2024chain,xie2024few}. As a result, they fall short in providing client-centered counseling~\citep{steenstra2024virtual} and systematic change talk exploration, especially given the inherent preference bias of LLMs~\citep{kang2024can}. Finally, the evaluation of counselor behavior in most works focus on comparing the agent generated responses with the predefined ``ground truth'' responses at the turn level~\citep{sun2024chain}. Such an evaluation approach rules out other possible high quality responses.  Meanwhile, human evaluation of counselor agents remains costly~\citep{xie2024few,hsu2023helping,steenstra2024virtual}. 
%This research task thus still lacks a more comprehensive evaluation framework.

\begin{figure*}
    \centering
    \includegraphics[width=\textwidth]{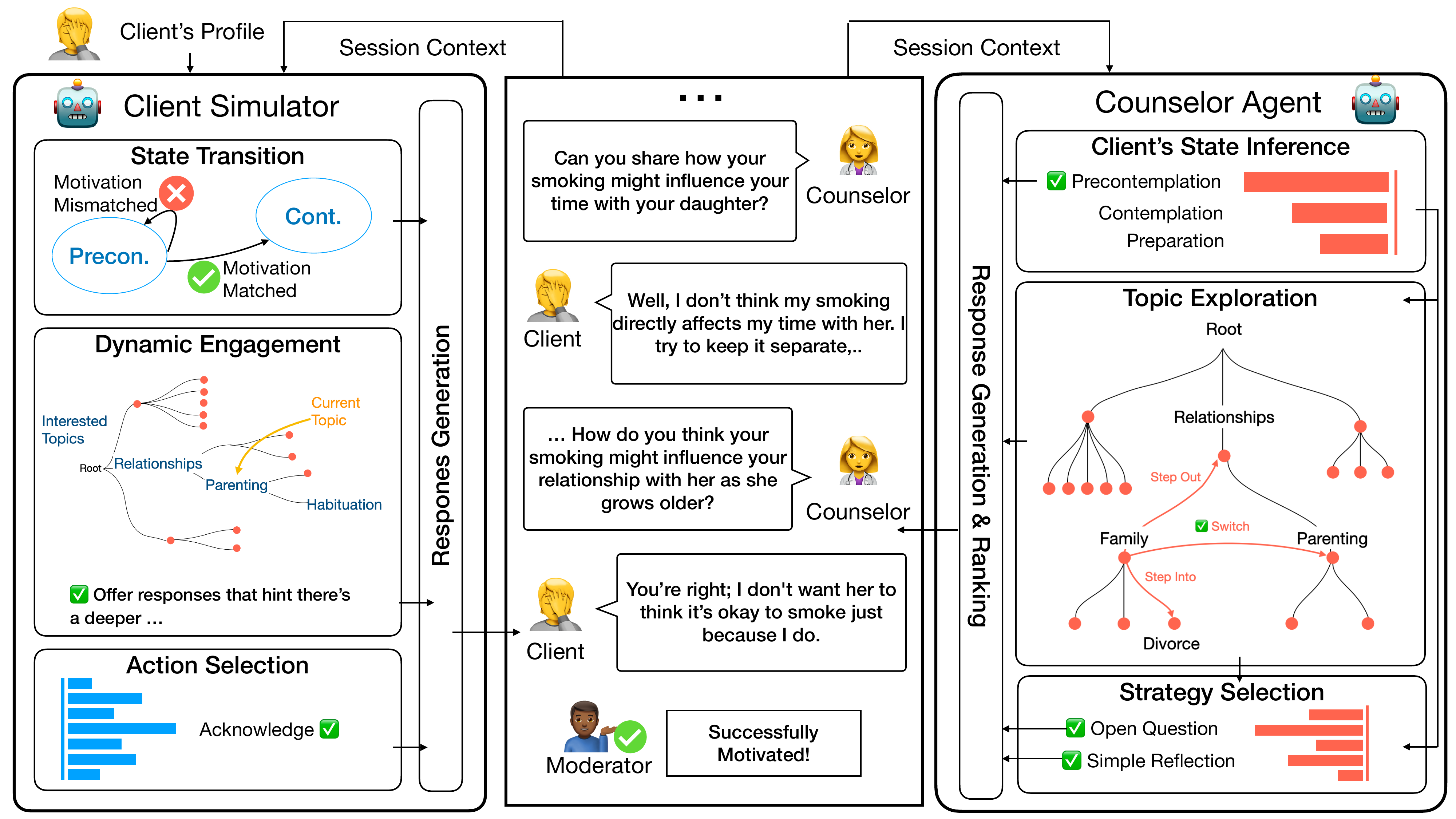}
    \caption{The STAR Framework illustrates how an LLM-based counselor agent and a client simulator can be created in an MI-based counseling session. 
    % The client simulator can be substituted with a human client if necessary. 
    The counselor agent infers the client's state, explores topics that motivate change talk, and utilizes MI techniques to generate appropriate responses. Check marks represent the selected actions, strategies, or states. The client agent may be simulated by adapting the STAR framework, incorporating modules for state transition, dynamic engagement, action selection, and response generation, allowing the client agent to closely align with the client's profile and adapt its engagement level with the counselor accordingly. The ``Precon.'' and ``Cont.'' indicate Precontemplation and Contemplation, respectively.}
    % \caption{Proposed STAR Framework. It shows how a LLM-based counselor agent and client agent can be simulated in a MI-based counseling session. The client agent can be replaced by a human client if required. The counselor agent infers the client's state, explores topics that trigger the motivation to change, and employs MI techniques to generate responses. We use check mark button to represented the selected action, strategy or state. For client simulation, STAR includes state transition, dynamic engagement, action selection, and response generation modules making the client agent follow the client's profile closely and adjust its engagement level. }
    \label{fig:framework}
\end{figure*}

{\bf Objective.} To tackle these challenges, we propose counselor agents that organically integrate psychotherapy skills, incorporating three novel ideas. First, we design a client’s state inference module that models the client's state of mind to provide personalized counseling tailored to different clients. This module draws inspiration from the transtheoretical model of health behavior change~\citep{prochaska2005transtheoretical,prochaska2008initial}, which defines five stages of change and suggests that the counselor should adjust their goals and strategies according to the client’s current stage. Second, we introduce a topic tree based on a real counseling dataset and a knowledge base (Wikipedia) that systematically models the topics of change talk exploration in MI counseling. With the help of the topic tree, the counselor can engage with a greater diversity of topics and focus more effectively on capturing client feedback to evoke change talk from clients with various inherent motivations. To the best of our knowledge, this is the first attempt to build a topic exploration method for a counselor agent. Third, we design a ranking-based approach for MI strategies guided response generation. Unlike previous approaches, which generate responses based on all selected strategies, we use a ranking approach for generated candidate responses for each strategy, effectively reducing the overuse of preferred strategies and mitigating bias in strategy selection. Finally, we develop a comprehensive automated evaluation framework using simulated clients that are carefully designed to exhibit realistic client behaviors.  The above ideas have been incorporated into our proposed framework known as the STAR (State, Topic, Action, and Response) framework as shown in Figure~\ref{fig:framework}. 
%In STAR, the evaluation of counseling agents covers multiple aspects of psychotherapy, such as MI skill competence and client experience. 
%We also incorporate professional MI experts to assess the performance of our counselor agent and framework. The experimental results demonstrate that our counselor agent effectively aligns with psychotherapy skills and adheres to MI principles, highlighting its potential for use in real-world MI applications.

{\bf Contributions.}  Our work contributes to the field of counselor agent research as follows: (1) Based on the STAR framework, we develop CAMI, a counselor agent that integrates state inference, topic exploration, and strategy selection modules, all aligned with established psychotherapy skills. (2) Through ablation study, we show that the ability of CAMI to infer client's state significantly improves its performance in MI-based counseling. (3) Our experiment also shows that topic tree and topic exploration further enhances CAMI's effectiveness in evoking change talk. 
%To the best of our knowledge, this represents the first attempt to create a counselor agent that simulates the MI counseling process as a professional human counselor would. 
(4) Our evaluation, combining both automated methods and human expert assessments, demonstrates that CAMI achieves superior performance, outperforming several strong baseline models. 
%\jjcomment{Should this paragraph be better aligned with the paragraph on ``Objective''? In the Objective paragraph, there were three novel ideas presented. It seems the three novel ideas are re-organized into item (1) and item (2) above. Shall we present them in consistent ways?}

\section{Related Work}
\label{sec:related}
%Motivational Interviewing (MI) is a client-centered counseling approach for eliciting behavior change by helping clients explore and resolve ambivalence~\citep{miller2012motivational,diclemente2002motivational,miller2002motivational,martins2009review}. 
Developing LLM-based agents to support mental health counseling is an emerging research topic. \citet{steenstra2024virtual} developed LLM-based agents for addressing the alcohol use behavior only. The counselor agent in DIIR~\citep{xie2024few} uses high-quality counseling sessions in the AnnoMI dataset~\citep{wu2022anno,wu2023creation} to learn a set of strategies in the form of natural language inductive rules. During counseling, DIIR uses the current session context to retrieve the best-matching strategy which prompts LLM to generate the next response. 
%Some works develop counselor agents using proprietary datasets to learn models for strategy prediction and generate the counselors' responses based on the predicted strategies. For example, 
\citet{sun2024chain} proposed Chain-of-Strategy (CoS), a prompting method, to align counselor response generation with MI strategies. CARE~\citep{hsu2023helping} determines the specific counseling strategy most suitable for a given context and provides example responses for peer counselors. However, the above works focus on modeling MI strategies only and %While strategy application is crucial for effective MI sessions, focusing %on essential counseling skills and adherence to MI principles, 
overlook the need to model the client's state and to evoke change talk based on topic(s) that could match the client's underlying motivation.  We thus propose to perform client's state inference and topic exploration on a topic tree specially constructed for MI counseling. %To the best of our knowledge, our work is the first attempt to consider topic exploration in eliciting change talk.

To compare the performance of counseling agents, the above works rely on evaluation approaches that compare the generated responses against the predefined ``ground truth'' responses.  One such approach computes word overlap or semantic similarity between the generated and ground truth responses.  Another approach requires human judgement comparing the generated responses with the ground truth ones at the turn-level when the counseling agent is given all previous turns following the ground truth session. 
As these approaches assume a single ``ground truth'' counseling session, they somewhat restrict the possibility of multiple high quality sessions thereby rating a competent counselor agent poorly only because it generates good responses very different from the ``ground truth.'' Unlike previous research, we utilize simulated clients representing a diverse range of behavioral issues to engage in motivational interviewing sessions with counselor agents. Using these simulated interactions, we conduct both automated and human evaluations to assess the performance of CAMI and other counselor agents.
\section{The STAR Framework}
\label{sec:counselor}
Our proposed STAR framework consists of four key modules that collaboratively mimic a counselor performing MI counseling, that is: (a) {\em \underline{S}tate Inference} (§\ref{sec: state estimation}), which infers the current client's state based on the transtheoretical model of health behavior change~\citep{prochaska2005transtheoretical,prochaska2008initial}. Based on the inferred client's state, the counselor may perform topic exploration before selecting a response strategy; (b) {\em \underline{T}opic Exploration} (§\ref{sec: topic exploration}), which explores a wide range of motivation topics in a topic tree to find the topic that may evoke change talk with the client; (c) {\em Strategy Selection}, also known as {\em \underline{A}ction Selection} (§\ref{sec: strategy selection}), which selects the appropriate strategy based on the current client's state, session context, and identified topic (if any); and (d) {\em \underline{R}esponse Generation \& Ranking} (see §\ref{sec: resopnse generation}), which generates a  few candidate responses based on the selected strategies and selects the most appropriate response that is coherent with the current context and aligned with MI principles.  In this study, we assume that the counselor is aware of the client's behavior problem before any counseling begins.

%{\color{blue} (EP: This para's content appears to duplicate the later sections)
%Right after a client's response, the counselor agent infers the client's state using the session context as input to a prompt submitted to the LLM. Then {\color{red} (EP: isn't topic exploration only done when the client is in pre-contemplation state?)} {\color{orange} (Yizhe: Yes, that is true. Thus we focus on the transition between precontemplaiton and contemplation state and end the sessions when the client is motivated.)}, the topic exploration module explores the topic tree and determines a topic that may be useful to evoke change talk. Simultaneously, the strategy selection module selects appropriate MI strategies to guide the counselor's response skillfully for. 
%To limit the response length, we restrict the number of selected strategies to fewer than two. The estimated state, along with the potential topic and strategies, are then used by the response generation module to generate the next response. Inspired by~\citet{hsu2023helping}, we generate one response for each selected strategy and for combined strategies, then rank these generated responses and choose the most coherent one to reply with. Details about the prompts can be found in the Appendix.
%} {\color{orange}(Yizhe: we can just remove it.)}

\subsection{Client's State Inference}
\label{sec: state estimation}

During MI counseling, a client may be in various mental states as they discuss their behavioral problems with the counselor. For counseling to be effective, the counselor must monitor the client's state and the current session context before responding ~\citep{prochaska2005transtheoretical,prochaska2008initial}. 

To define the client's states~\citep{prochaska1997transtheoretical, hashemzadeh2019transtheoretical}, we utilize the transtheoretical model of health behavior change~\citep{prochaska2005transtheoretical,prochaska2008initial}, consisting of five states: {\em Precontemplation}, {\em Contemplation}, {\em Preparation}, {\em Action}, and {\em Maintenance}.  In the {\em Precontemplation} stage, the client may not seriously consider behavior change nor be motivated to seek help. In the {\em Contemplation} stage, the client may weigh the pros and cons of changing their behavior. In the {\em Preparation} stage, the client has made a commitment to change. In the {\em Action} stage, the client actively takes steps to change their behaviors. In the {\em Maintenance}, the client seeks to avoid any temptation to return to the bad behavior. 
%we only consider the first three states, as MI counseling is particularly useful when a client is in the Precontemplation or Contemplation states. The 

Our primary goal is to assist clients in reaching the Preparation state, so our counselor agent focuses on the first three states. In the Precontemplation state, the counselor focuses on building trust, establishing change goals, understanding the client's motivations, and addressing misconceptions before encouraging the client to change. In the Contemplation state, the counselor listens to the client and addresses incorrect beliefs. In the Preparation state, the counselor provides the necessary information and tools to guide the discussion. 
We also aim to equip the counselor with the ability to %engage the client and evoke behavior change. We thus focus on the transition from Precontemplation to Contemplation where the counselor seeks to 
engage the client, to collaborate with the client to identify the motivating topic for change, and to soften the client's preference for the status quo.
%Given their strong capabilities in language understanding, classification, and generation~\citep{ouyang2022training,achiam2023gpt,dubey2024llama,brown2020language}, 

We thus prompt LLMs to infer the client's state based on the session context and state descriptions in a zero-shot manner. Our experiment results in §\ref{sec:topic exploration} show that the state-of-the-art LLMs perform this state inference task with high accuracy (see §~\ref{sec:state_inf_accuracy}).
\subsection{Topic Exploration}
\label{sec: topic exploration}

When the client is in the Precontemplation state, the counselor engages the client to develop rapport and cultivate change talk by supporting them in exploring topics that evoke motivation for behavior change. During the process, the counselor may explore different topics (e.g., smoking causing harm to family members, medication reducing cholesterol)~\citep{cole2023using} also known as the motivation topics. 
%While the current LLM-based agents are abundant in knowledge, they may have preference biases~\citep{kang2024can}{\color{red} (EP: topic bias?)}, which limit the diversity of change talk exploration. {\color{red} (EP: I thought this bias refers to bias in strategies, and not about topic exploration. Our work should be the first to introduce topic exploration.)} 
To our knowledge, topic exploration has not been studied in previous counseling agent research~\citep{hsu2023helping,xie2024few,sun2024chain} as most methods ignore the importance of change talk exploration. 

\noindent
\textbf{Topic Tree Construction.} We propose a three-level topic tree as an external knowledge base that facilitates the exploration of motivation topics by the counselor agent, aiming to evoke change talk for clients facing various behavioral challenges. As shown in Figure~\ref{fig:topic tree}, the topic tree is structured into superclass topics, coarse-grained topics, and fine-grained topics. To ensure standardization, we require all topic labels to match Wikipedia article titles, ensuring each topic is represented by a dedicated page. The construction of the topic tree draws on both real counseling sessions and the capabilities of GPT-4o, following this methodology:

%\begin{enumerate} 
%    \renewcommand{\labelenumi}{Step~\theenumi.} 
    %\item 
    \noindent
    \textit{Step 1: Derive Fine-Grained Topics.} Firstly, we derive a set of fine-grained topics from some counseling session corpus. Ideally, we would extract topics from a larger corpus to ensure comprehensive coverage. However, due to data limitations, we utilize the available AnnoMI dataset~\citep{wu2022anno,wu2023creation} and identify 28 fine-grained topics. 
    
    \noindent
    \textit{Step 2: Group Coarse-grained and Fine-grained Topics.} Next, we organize the derived fine-grained topics into coarse-grained topics based on Wikipedia's parent-child category relationships. Similarly, the coarse-grained topics are grouped into superclass topics.
    %, we group the fine-grained topics under the same category as an abstract topic.  To better model the hierarchical structure of topics, we create two levels of abstract topics. 
    Finally, we construct three levels of the topic tree: superclass topics (e.g., {\em Health} and {\em Economy}), coarse-grained topics (e.g., {\em Disease} and {\em Mental Disorders}), and fine-grained topics (e.g., {\em Depression} and {\em Hypertension}). 
    %All fine-grained topics can be linked to a Wikipedia page, and each abstract topic can be linked to a Wikipedia category. {\color{red} (EP: Do we have an article for each superclass and coarse-grained topic?)}
    
    \noindent 
    \textit{Step 3: Expand Topic Tree.} To broaden the range of topics, we expand the initial topic tree by prompting GPT-4o to generate additional coarse- and fine-grained topics, leveraging its extensive topical knowledge. We choose not to expand the superclass topics as they are already very general. Instead, we prompt GPT-4o to suggest topics under the superclass and coarse-grained topics in a few-shot manner (see Table~\ref{tab:topic expansion}). This step continues until there are no more new topics. Our final topic tree consists of 5 superclass topics, 14 coarse-grained topics, and 59 fine-grained topics, all verified as titles of Wikipedia pages, as shown in Figure~\ref{fig:topic tree}.
%\end{enumerate}

\noindent
\textbf{Topic Navigation.} We guide the counselor agent in navigating topics within the topic tree to initiate change talk in two stages. 
%Specifically, at each turn, the counselor takes in the client's feedback to the current topic. 
In the {\em initial engagement stage}, the counselor agent is provided with the superclass topics and prompted to %interact with the client based on these topics, aiming to 
explore these topics broadly with the client. The counselor agent should infer the client's interested topic (i.e., the one that evokes change talk) and estimate the probability of each superclass topic being the interested topic in each turn (see the prompt in Table~\ref{tab:topic initialization}). This phase ends after six turns which is similar to the number of turns for initial engagement of client proposed in~\citet{park2019designing}, or when a topic is assigned a probability higher than 40\%. The counselor will select the topic with the highest probability as the \textit{current client topic (or current topic)}.

In the subsequent {\em focused engagement stage}, when the inferred client topic does not match the client's actual motivation topic, the counselor agent performs one of the three navigation operations to identify other possible topics in the topic tree: (a) {\em Step Into}: This operation explores a sub-topic (either coarse-grained or fine-grained) of the current topic when the client provides some positive feedback, indicating a desire for a deeper discussion. The LLM is prompted to select the next topic from all sub-topics based on previous context and exploration path; (b) {\em Switch}: This operation allows the counselor to switch to another topic at the same level (i.e., sharing the same parent topic as the current topic) when the client shows limited engagement with the current topic but expresses interest in the parent topic. Similar to Step Into, the LLM is prompted to select the next topic from all candidate topics; and (c) {\em Step Out}: This operation involves revisiting the parent topic when the client does not wish to continue discussing the current topic or the parent topic. This occurs when the counselor has navigated to the wrong sub-tree. The prompt for selecting the navigation operations is shown in Table~\ref{tab:topic exploration}.
%In each operation, all the candidate topics are provided in instruction for selection by the agent.
%Based on the previous action, topic exploration module focuses on a specific topic according to the previous context, particularly considering client feedback, which guides the counselor's response generation. 
All the above navigation operations are facilitated by prompting the LLMs as shown in Appendix~\ref{app:counselor implementation}.  An example of topic exploration is also shown in Appendix~\ref{app:example}.

\subsection{Strategy Selection}
\label{sec: strategy selection}

Previous research has shown the importance of strategy selection in effective AI-based counseling~\citep{kang2024can,xie2024few,hsu2023helping,sun2024chain}. In our STAR framework, the Strategy Selection module select strategies from the Motivational Interviewing Skill Code (MISC)~\citep{miller2003manual} to guide LLMs in demonstrating MI skills, thereby aligning the LLMs with strategies that offer controllability and explainability. 
 Inspired by~\citet{sun2024chain}, we employ zero-shot prompts providing the definitions of different MI strategies~\citep{miller2003manual,miller2012motivational} and the current inferred client's state. A LLM prompt is then used to select MI strategies to guide response generation. 
 Both prior works~\citep{sun2024chain,xie2024few} and our own MI competency evaluation results (see Table~\ref{tab:miti}) highlight the effectiveness of this method.
 % Previous works~\citep{sun2024chain,xie2024few} and our experiment results in MI competency (Table~\ref{tab:miti}) demonstrate the effectiveness of this method through assessing the final generated response.

Following recent research~\citep{sun2024chain,xie2024few}, we allow the strategy selection module to select multiple strategies for generating a response (see Table~\ref{tab:strategy selection}). However, we limit the number of selected strategies to two so as to avoid an excessive number of strategies (e.g., more than five) in one generated counselor response. This limit is consistent with what we observe in the AnnoMI~\citep{wu2022anno,wu2023creation} dataset. 

\subsection{Response Generation \& Ranking}
\label{sec: resopnse generation}

Finally, the response generation module generates the counselor's response in the next turn using a {\em turn-by-turn} generation approach. 
Specifically, we construct an instruction prompt for the MI counselor, which includes the client's behavioral issue, the goal to achieve, the session context, and a turn-level instruction. The turn-level instruction further includes a description of the inferred client state, the current client topic, and the selected strategy(ies) (see Table~\ref{tab:couneslor system}).

When two strategies are selected, we follow the approach in \citet{hsu2023helping}, which generates an response for each single strategy as well as for the concatenated selected strategies. After generating the three candidate responses, we prompt the LLM to choose the most coherent  response based solely on the current session context (see Table~\ref{tab:response rank}).

\section{Experiments}
%Our experiment focuses on evaluating the counselor agent's ability to demonstrate MI competency and counseling effectiveness. First, we design the experiment setup involving simulated clients consulting different counseling agents using our proposed counselor simulation method and other baseline methods (See §\ref{sec:experiment setting}). For fair comparison, we adopt only one client simulation method which programs a LLM to mimic clients with different  profiles (i.e., behavior problems, motivation topics, and personal attributes).  We propose an evaluation scheme covering three criteria, namely: (a) MI competency, (b) client experience, (c) success in motivating change, and (c) topic exploration.  We finally report the counselor agent's performance in Section~\ref{sec:mi competence}, \ref{sec:client evaluation}), and \ref{sec:topic exploration}) respectively. Finally, we report the results of a human expert evaluation (§\ref{sec:expert evalution}) comparing counselor agents using our method and COS against the human counsellors.

\subsection{Experiment Setup}
\label{sec:experiment setting}

%\subsubsection{Generation Framework}
Following the STAR framework and experiment setup used in previous works~\citep{tu2024towards,chiu2024computational,yosef2024assessing,wu2023towards}, we implement three specialized LLM-based agents, namely, {\em client}, {\em counselor}, and {\em moderator} agents. Their respective prompts can be found in the Appendix~\ref{app:implementation}.

The counselor agent is role-played using some counselor agent methods. Other than CAMI, we include other baseline methods described in §\ref{sec:baselines}.  The client agent is also simulated using different input profiles. To ensure the client's behavior is realistic and consistent with real users, we develop the client agent to maintain its state and generate responses consistent with its state, profile, and current session context using multiple modules instead of a simple prompt used in previous works~\cite{chiu2024computational,yosef2024assessing,wu2023towards}. The detailed client simulation design is given in Appendix~\ref{app:client simulation}. 
The moderator oversees the conversation session between the counselor and client agents, and decides to end the session when: 1) the counselor or client proactively concludes the session, 2) the client becomes motivated to change, or 3) the session reaches the maximum of 50 turns (which is similar what we observed in real world sessions, e.g., AnnoMI).

We employ \texttt{gpt-4o-2024-08-06} as the LLM backbone for both the client and the moderator. For counselor, we employ both \texttt{gpt-4o-2024-08-06} and \texttt{Llama-3.1-70B} as the LLM backbone.
The session simulation begins with the counselor's response: ``\textit{Hello. How are you?}'' The client then responds with ``\textit{I am good. What about you?}'' Based on the session history, the counselor continues to generate the next response. This process repeats until the moderator decides to end the session.

\subsection{Baseline Counselor Simulation Methods}
\label{sec:baselines}

%DIIR~\citep{xie2024few} is a retrieval-augmented framework that infers natural language strategies from expert demonstrations and reuses these strategies at inference time. 
We compare our counselor agent method with four state-of-the-art LLM-based counselor agent methods, including: 1) \textbf{Base}: Following \citet{steenstra2024virtual}, the Base counselor agent incorporates only a system prompt that includes the task description and knowledge about MI techniques. 2) \textit{Dialogue Strategy Learning via Inductive Reasoning} (\textbf{DIIR}) (see §~\ref{sec:related}) 
3) \textit{Chain of Strategy} (\textbf{CoS}): Following \citet{hsu2023helping,sun2024chain}, the CoS (Chain-of-Strategy) counselor agent uses a chain of modules to generate responses after strategy selection. Since the original CoS prompt is not available, we implement it as an ablated version of CAMI without state inference and topic exploration. 4) \textit{CAMI without Topic Exploration} (\textbf{CAMI-TE}) To evaluate the effectiveness of topic exploration, we introduce CAMI-TE, which is the ablated version of CAMI without the topic exploration module. 
%{\color{red} (EP: Does the retrieval-augmented framework improve our method?)} {\color{orange} (Yizhe We haven't incorporated this method in our framework as we don't focus on strategy selection before.)} 
To prevent data leakage, we use other sessions in AnnoMI~\citep{wu2022anno,wu2023creation} not included in the client's set as demonstrations.

We use the high-quality subset of the AnnoMI dataset~\citep{wu2022anno,wu2023creation} and manually annotate 38 client profiles whose sessions contain sufficient information and were successfully motivated by human counselors. These clients cover a good diverse set of motivation topics and interests. To obtain more reliable results, each counselor simulation method generates five sessions for each client profile. We therefore have 190 (= 38 $\times$ 5) generated sessions for each method.  For the automated evaluation that follows, we average the results for the 5 sessions generated for each client profile before averaging over all the 38 client profiles to derive the results of each counselor simulation method. 

\subsection{MI Competency}
\label{sec:mi competence}
The Motivational Interviewing Treatment Integrity (MITI)~\citep{moyers2016motivational} and Motivational Interviewing Skill Code (MISC)~\citep{miller2003manual} are commonly used to evaluate the competency of MI counselors. 

\noindent
\textbf{MITI Behavior Scores.} Following~\citet{xie2024few}, we employ a fine-tuned classifier on crowd-sourced dataset~\citep{welivita2022curating} to retrieve the five most possible behavior labels, followed by GPT-4o deciding the final label out of the five. The detailed implementation can be found in Appendix~\ref{app:miti annotation}.  We report the following behavior-counts: 1) \emph{Reflection Question Ratio (R/Q)}, defined by the number of reflections by the counselor agent divided by the number of questions asked by the counselor; 2) \emph{Proportion of Open Questions (\%OQ)}, defined by the number of open questions asked by counselor divided by the number of questions (including both open and closed questions); 3) \emph{Proportion of Complex Reflections (\%CR)}, defined by the number of paraphrase and summary reflections divided by the total number of reflections; 4) \emph{Proportion of MI-Consistent Responses (\%MIC)}, defined by the number of MI-consistent responses divided by the sum of MI-consistent and MI-inconsistent responses; and 5) \emph{Percent Therapist Talk Time (\%TTT)}, defined by the number of minutes the counselor spoke divided by the total talk time (counselor and client combined). Word count is used to approximate talk time instead.

\begin{table}[tb]
\resizebox{0.49\textwidth}{!}{
\begin{tabular}{clllll}
\toprule
                      & R/Q$\uparrow$    & \%OQ$\uparrow$    & \%CR$\uparrow$    & \%MIC$\uparrow$   & \%TTT$\downarrow$          \\ \midrule
Expert                & \textgreater 2.0 & \textgreater 70\% & \textgreater 50\% & \textgreater 90\% & \textless 50\% \\
Prof.           & \textgreater 1.0 & \textgreater 50\% & \textgreater 40\% & \textgreater 80\% & \textless 60\% \\ \midrule
HQ                  & 1.28             & 52.1\%$^p$       & 51.4\%$^e$           & 82.4\%$^p$           & 53.4\%$^p$ \\ 
LQ                   & 0.34             & 16.2\%           & 14.22\%           & 40.5\%           & 59.6\%$^p$\\ \midrule
\multicolumn{6}{c}{\cellcolor[HTML]{E0E0E0}GPT-4o Based Counselor}                                                           \\
Base                  & 0.18             & 92.3\%$^e$             & 37.6\%            & 85.0\%$^p$            & 42.7\%$^e$         \\
DIIR                  & 0.42             & 98.9\%$^e$             & \textbf{82.4}\%$^e$   & 89.1\%$^p$            & 54.9\%$^p$         \\
CoS                   & 0.29             & 96.5\%$^e$             & 49.1\%$^p$            & 94.5\%$^e$            & 38.3\%$^e$         \\
C$\ast$-TE                & 0.37             & 97.2\%$^e$             & 51.1\%$^e$            & 94.7\%$^e$            & 36.1\%$^e$         \\
CAMI                  & \textbf{0.56}    & \textbf{99.0}\%$^e$    & 57.3\%$^e$            & \textbf{96.6}\%$^e$   & \textbf{34.8}\%$^e$         \\ \midrule
\multicolumn{6}{c}{\cellcolor[HTML]{E0E0E0}Llama-3.1 70B Based Counselor}                                                    \\
Base                  & 0.34             & 34.3\%            & 13.1\%            & 55.4\%            & 47.9\%$^e$         \\
DIIR                  & 0.97             & 76.8\%$^p$           & \textbf{71.5}\%$^e$   & 85.4\%$^p$            & 45.0\%$^e$         \\
CoS                   & 0.77             & 75.2\%$^p$           & 58.9\%$^e$            & 87.1\%$^p$            & 43.9\%$^e$         \\
C$\ast$-TE  & 0.89             & 76.6\%$^p$           & 61.2\%$^e$            & 88.9\%$^p$            & 40.4\%$^e$         \\
CAMI                  & \textbf{1.11}$^p$  & \textbf{83.7}\%$^e$   & 60.5\%$^e$            & \textbf{90.7}\%$^e$  & \textbf{38.4}\%$^e$         \\ \bottomrule
\end{tabular}}
\caption{MI behavior count-based evaluation results. ``Prof.'' and ``C$\ast$-TE'' denote Proficiency and CAMI-TE respectively.  CAMI outperforms other baselines in most metrics, except for the percentage of complex reflections. Although DIIR shows a high percentage of complex reflections, it may overuse them, which is also a concern~\citep{steenstra2024virtual}. All models perform poorly on the R/Q metric, indicating a significant preference for questions over reflections. The values with $^e$ and $^p$ superscripts meet the thresholds for ``Expert'' and ``Proficiency'' levels respectively. HQ and LQ represent the performance of human counselors in high- and low-quality AnnoMI sessions respectively.}
\label{tab:miti}
\end{table}

\noindent
\textbf{MITI Global Scores.} We also report four MITI global scores, assigned on a 5-point Likert scale for each counseling session, including: 1) \emph{Cultivating Change Talk}, which measures the extent to which the clinician actively encourages the client’s language in support of the change goal and their confidence in making that change; 2) \emph{Softening Sustain Talk}, which measures the extent to which the clinician avoids focusing on reasons against change or for maintaining the status quo; 3) \emph{Partnership}, which measures the extent to which the clinician conveys an understanding that expertise and wisdom about change primarily reside within the client; and 4) \emph{Empathy}, which measures the extent to which the clinician attempts to understand the client’s perspective and experience, essentially trying to "try on" what the client feels or thinks. The global score prediction task is designed to take the entire session transcript as input and predict a single value (1-5) for one of the four global scores. We employ GPT-4o with the prompt from previous work~\citep{cohen2024motivational}.

\begin{table}[tb]
\resizebox{0.49\textwidth}{!}{
\begin{tabular}{lrrrr}
\toprule
     & Cultivate$\uparrow$ & Soften$\uparrow$ & Partner$\uparrow$ & Empathy$\uparrow$ \\ \midrule
HQ & 2.85                    & 2.88                   & 2.97        & 3.49    \\
LQ  & 2.43                    & 2.47                   & 2.26        & 2.45    \\ \midrule
\multicolumn{5}{c}{\cellcolor[HTML]{E0E0E0}GPT-4o Based Counselor}                     \\
Base & 2.33                    & 2.70                   & 2.15        & 3.04    \\
DIIR & 2.21                    & 2.64                   & 2.37        & 3.10    \\
CoS  & 2.40                    & 2.71                   & 2.33        & 3.23    \\
CAMI-TE  & 2.43                 & 2.78                   & 2.37        & 3.28    \\
CAMI & \textbf{2.62}           & \textbf{2.86}          & \textbf{2.58} & \textbf{3.37}    \\ \midrule
\multicolumn{5}{c}{\cellcolor[HTML]{E0E0E0}Llama-3.1 70B Based Counselor}              \\
Base & 2.10                    & 2.34                   & 2.08        & 2.57    \\
DIIR & 2.13                    & 2.57                   & 2.28        & 2.87    \\
CoS  & 2.14                    & 2.61                   & 2.13        & 3.07    \\
CAMI-TE  & 2.20                 & 2.64                   & 2.23        & 3.12    \\
CAMI & \textbf{2.38}           & \textbf{2.78}          & \textbf{2.37} & \textbf{3.33}    \\ \bottomrule
\end{tabular}}
\caption{Results of the MI global score evaluation. Our method outperforms others in all four scores. Note that ``Cultivate'', ``Soften'' and ``Partner'' are abbreviations for Cultivating Change Talk, Softening Sustain Talk, and Partnership respectively. }
\label{tab:global score}
\end{table}

As shown in Tables~\ref{tab:miti} and ~\ref{tab:global score}, CAMI outperforms the baselines (including DIIR) in most MI competency metrics. This suggests that CAMI is capable of selecting MI-adherent strategies during counseling. While DIIR outperforms our method in \%CR, the very high \%CR of DIIR (82.4\%) may indicate overuse of complex reflection~\citep{steenstra2024virtual}. On the other hand, all counselor agents perform poorly on R/Q, indicating a preference for using questions instead of reflection (especially open questions, as indicated by the high \%OQ). Upon manual inspection of the generated conversations, we found that almost every counselor's response involved raising questions to continue the session, which differs from the human counselors in the HQ sessions.

\subsection{Success Rate and Exploration Focus}
% \subsection{Success Rate in Outcome and Topic Exploration}
\label{sec:topic exploration}

\textbf{Success rate.} Successful change talk is the main goal of MI. We therefore introduce \textit{success rate} defined by the proportion of clients with whom the counselor agent successfully evokes change talk, resulting in increased motivation to change at the end of session. Since successful evocation of change talk occurs when the counselor and client collaboratively identify the underlying motivational topics, we also evaluate success rates for each superclass topic. As shown in Table~\ref{tab:success}, CAMI demonstrates the highest success rate in all five superclass topics, achieving the best overall performance. In contrast, counselor agents without topic exploration (i.e., Base, DIIR, CoS and CAMI-TE) tend to focus on only a few topics, such as Health and Relationships, and often fail to explore the less common topics (e.g., Law, Education, and, to some extent, Economy). %Additionally, we observe variation in topic preferences across different LLMs; for example, the Llama-based counselor tends to raise economy-related topics, whereas the GPT-4o-based counselor does not.

\begin{table}[tb]
\resizebox{0.49\textwidth}{!}{
\begin{tabular}{crrrrr|c}
\toprule
 & HE & EC & RE & LA & ED & Overall \\ 
 & (105) & (30)& (40)& (5)& (10)& (190)\\ \midrule
\multicolumn{7}{c}{\cellcolor[HTML]{E0DEDE}GPT-4o Based Counselor}                                                                                                   \\
Base      & 47.6 & 0.0   & 25.0                                                                & 0.0 & 0.0     & 31.5                    \\
DIIR      & 52.4 & 0.0   & 27.5                                                                & 0.0 & 0.0     & 34.7                    \\
CoS       & 54.3 & 0.0   & 32.5                                                                & 0.0 & 20.0    & 37.9                    \\
CAMI-TE    & 54.3 & 0.0   & 32.5                                                                & 0.0 & 30.0    & 38.4                    \\
CAMI      & \textbf{57.1} & \textbf{23.3}  & \textbf{70.0}     & \textbf{40.0} & \textbf{40.0}    & \textbf{53.1}                    \\ \midrule
\multicolumn{7}{c}{\cellcolor[HTML]{E0DEDE}Llama-3.1 70B Based Counselor}                                                                                            \\
Base      & 48.6 & 6.7   & 12.5                                                                & 0.0 & 0.0     & 30.5                    \\
DIIR      & 57.4 & 10.0  & 10.0                                                                & 0.0 & 0.0     & 35.2                    \\
CoS       & 60.0 & 26.6  & 22.5                                                                & 0.0 & 0.0     & 42.1                    \\
CAMI-TE    & 60.0 & 30.0  & 22.5                                                                & 0.0 & 0.0     & 42.6                    \\
CAMI      & \textbf{61.9} & \textbf{36.7}  & \textbf{45.0}     & \textbf{40.0}  & \textbf{20.0}   & \textbf{51.1}              \\ \bottomrule
\end{tabular}}
\caption{Success rate of counselor agents for clients with different superclasses of motivation topics (HE: Health, EC: Economy, RE: Relationship, LA: Law, ED: Education). The number of clients with motivation topics covered by each superclass is shown in parentheses. CAMI achieves highest overall success rate, and is able to cope well in all superclass topics including the less popular ones.}
\label{tab:success}
\end{table}

{\bf Exploration focus.} To further demonstrate the impact of topic exploration,
%enhances the counselor agent's ability to explore a wider range of topics and delve deeper into specific topics, we measure \textit{exploration diversity} and \textit{exploitation precision}. Exploration diversity is assessed through Shannon Entropy, which measures the breadth of topics (referred to as diversity) the counselor engages with across all generated conversations. For this, GPT-4o is utilized to categorize each counselor utterance into topics based on our predefined topic tree structure, focusing only on superclass and coarse-grained topics. 
we use GPT-4o to assess how close are the topics explored in counselor's responses from the client's fine-grained motivation topic as counselor agents using baseline methods could not explore topics.  We thus prompt GPT-4o to take a counselor's response and client's fine-grained topic as input, determine the topic embedded in the former, and return the exploration focus metric.  Exploration focus (FOC) is a number from 0 (exact match between the topic embedded in counselor's response and client's topic at the fine-grained level) to 3 (the topic embedded in counselor's response does not match the superclass, coarse-grained and fine-grained topics of the client). As shown in Table~\ref{tab:diversity}, CAMI excels at exploration focus which gives it the edge to evoke change talk.
%delving into specific topics by perceiving client feedback. The diverse topic coverage indicates the counselor's capacity to robustly engage with different clients, while closer alignment to the ground truth topic suggests the agent's ability to effectively evoke change talk.

As CAMI can reveal its explored topics, we also evaluate how well the client agent can correctly decipher the topic embedded in the counselor's response. Our results show that the client is quite accurate in determining the correct explored topic 74.92\% and 70.14\% of the time with CAMI using GPT-4o and Llama-3.1 70B respectively.

\begin{table}[tb]
\centering
\scalebox{0.8}{
\begin{tabular}{l|ccccc}
\toprule
              & Base & DIIR & CoS  & CAMI-TE & CAMI \\ \midrule
GPT-4o        & 2.62 & 2.59 & 2.64 & 2.60    & \textbf{2.37} \\
Llama-3.1 70B & 2.81 & 2.72 & 2.77 & 2.74    & \textbf{2.44} \\
\bottomrule
\end{tabular}
}
\caption{Exploration Focus Results ($\downarrow$).}
\label{tab:diversity}
\end{table}

\subsection{State Inference Accuracy}
\label{sec:state_inf_accuracy}

We evaluate the accuracy of inferring client's state by the state inference module of the CAMI counselor agent.  Since CAMI-TE shares the same state inference module, it shares the same state inference accuracy as CAMI. We could not perform the same evaluation on the other baseline methods as they are not equipped with the state inference module. Our experiments show that CAMI using GPT-4o and Llama-3.1 70B infer the correct client state most of the time at 93.32\% and 88.56\% respectively.  This high accuracy thus allows CAMI to generate more client-centered responses, increasing the effectiveness of MI counseling.

\subsection{Expert Evaluation}
\label{sec:expert evalution}

In addition to automatic evaluation, we also involve three co-authors (two with strong counseling credential and expertise and another is a social science student with rich experience in peer support and counseling) to assess the overall performance of CoS and CAMI. Each assessor is to rate the simulated counselor counseling clients simulated using five client profiles randomly selected from AnnoMI such that both the high and low quality sessions for these clients exist in AnnoMI. The sessions are then evaluated by MITI Ratings, Change Talk Exploration, Success in Eliciting Change Talk, and Realism of both Counselor and Client. \footnote{Detailed instructions about these measures are given in Appendix~\ref{app:expert evaluation}.}

As shown in Table~\ref{tab:expert evaluation}, other than the counselors in HQ sessions, CAMI yields the best results across all the metrics, significantly outperforming the CoS counselor (p-value $< 0.005$). The counselors in LQ sessions unsurprisingly perform the worst. Although CAMI performs better than other LLM-based counselor agent, there is still room for improvement compared to high-quality human counselors. Experts also provided feedback regarding CAMI's performance, noting some behaviors that are not recommended, such as overusing questions and consistently reflecting on the positive aspects of the client's problematic behavior, indicating areas for future improvement.

\begin{table}[tb]
\resizebox{0.49\textwidth}{!}{
\begin{tabular}{cc|cc|cc}
\toprule
\multicolumn{2}{c|}{}                                             & HQ        & LQ        & CoS  & CAMI \\ \midrule
\multicolumn{1}{r|}{\multirow{4}{1.5cm}{MITI Global Ratings}} & Cultivating $\cdots$ & 4.06      & 1.68      & 2.74 & \textbf{3.68} \\
\multicolumn{1}{r|}{}                             & Softening $\cdots$  & 3.90      & 1.74      & 2.74 & \textbf{3.32} \\
\multicolumn{1}{r|}{}                             & Partnership  & 4.26      & 1.46      & 3.60 & \textbf{3.90} \\
\multicolumn{1}{r|}{}                             & Empathy      & 4.26      & 1.38      & 3.72 & \textbf{4.00} \\ \midrule
\multicolumn{2}{r|}{Change Talk Exploration}                      & 4.18      & 1.58      & 3.00 & \textbf{3.94} \\ \midrule
\multicolumn{2}{r|}{Evoking Change Talk}                        & 2.68      & 1.20      & 1.80 & \textbf{2.40} \\ \midrule
\multicolumn{2}{r|}{Realism of Counselor}                            & 4.68      & 2.32      & 3.06 & \textbf{3.60} \\ \midrule
\multicolumn{2}{r|}{Realism of Client}                               & 4.68      & 4.00      & 4.00 & \textbf{4.32} \\ 
\bottomrule
\end{tabular}}
\caption{Results of expert evaluation. ``Cultivating $\cdots$'' and ``Softening $\cdots$'' are abbreviations for "Cultivating Change Talk" and "Softening Sustain Talk" respectively. CAMI outperforms CoS and LQ counselors but a gap still exists between CAMI and HQ counselors. Note that the maximum score of ``Evoking Change Talk'' is 3, while that of other criteria is 5.}
\label{tab:expert evaluation}
\end{table}

\section{Conclusion}
In this paper, we address the challenges of using LLMs to create counselor agents that practice motivational interviewing style counseling using our proposed STAR framework.  This framework includes client's state inference, topic exploration for change talk, and MI-adherent strategy selection.  Based on the framework, we develop CAMI a counselor agent that engages clients in change talks making positive changes to their behaviors. By extracting motivation topics and their superclass relationships from AnnoMI dataset and Wikipedia knowledge base, we construct a novel topic tree covering hierarchical motivation topics that can be navigated by CAMI as it explores the topics of change talk for clients having diverse profiles. Furthermore, CAMI selects strategies from MI skills to generate strategy-aligned responses. Extensive experiments and analyses, including automated and expert user evaluations, show that the STAR framework underlying CAMI effectively guides LLMs to adhere to MI techniques. CAMI also outperforms state-of-the-art counseling agents in our comprehensive evaluation, reducing the gap between human and AI-based counseling. 

\section*{Limitations}
Although CAMI shows promising results in MI-based counseling, we could identify three main limitations which can be addressed in the future work.

Firstly, our work has not studied how sensitive is CAMI to the multiple steps of prompting due to the accumulation of generation errors. Even though some aspects of the performance, such as topic exploration, are not optimal, we believe that other methods could be implemented to improve module performance while maintaining the novelty and effectiveness of the overall framework. Designing effective prompts and evaluating their performance are therefore important areas for further research. 

Secondly, our framework currently relies on LLM prompting to perform state inference, topic exploration and strategy selection. Our human experts have raised concerns about issues related to strategy use, such as counselor responses being too lengthy and using too many questions. In our future work, we would therefore plan to enhance the strategy selection module further such as incorporating the idea of action rules introduced in DIIR. 

Finally, our study focuses on MI-based counseling and the AnnoMI dataset.  However, the proposed STAR framework can be adapted to other counseling techniques, such as Cognitive Behavioral Therapy (CBT), and relevant counseling datasets.  Future research could explore the integration of MI and CBT techniques within a single counselor agent, potentially enhancing the counseling effectiveness. 

\section*{Ethical Considerations}
This work does not advocate for the direct use of LLMs in therapy. Although our proposed CAMI system outperforms previous methods and demonstrates proficiency in MI techniques, it still falls short of the expertise provided by high-quality human counselors. Therefore, we do not suggest that LLMs should replace the role of human counselors. While the outputs of LLMs are certainly not as specialized or accurate as those of professionals, they can serve as supplementary tools to support and inspire human-led exploration. Our goal is to provide an alternative approach for applying LLMs in psychological support, offering ideas that may spark further discussions and development in the field of NLP and psychology. We believe this can help advance future research in AI applications for psychology and sociology. Moreover, while MI techniques aim to facilitate positive change, system deployers should carefully assess the topics covered to mitigate potential risks. Additionally, users interacting with systems incorporating MI abilities should be informed of the system's identity, ensuring they are aware of the nature of their interaction and can make an informed choice.

\bibliography{custom}

\appendix

\section{Detailed Implementation}
\label{app:implementation}

In this section, we provide the detailed implementation of our method and experiment evaluation, which includes the prompts and instructions for human evaluation. We set the top-p and temperature of to 0.7 and 0.8 for LLMs.

\subsection{Counselor Agent Implementation}
\label{app:counselor implementation}

\begin{figure*}
    \centering
    \includegraphics[width=\textwidth]{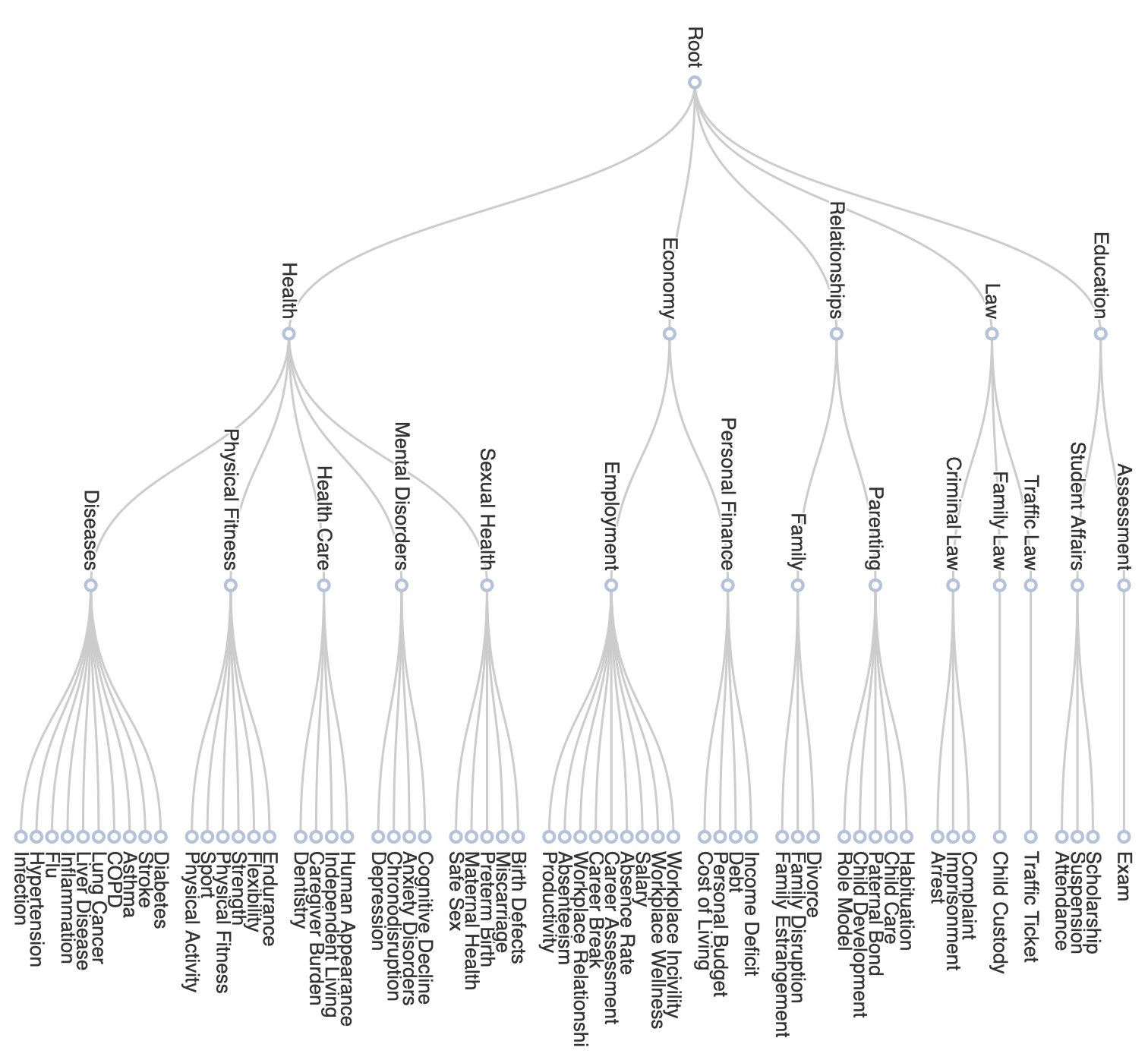}
    \caption{The topics tree constructed in our work consists of 5 Super-Class topics (i.e., {\em Health}, {\em Economy}, {\em Relationship}, {\em Law}, and {\em Education}), 14 Coarse-Grained Topics, and 59 Fine-Grained Topics.}
    \label{fig:topic tree}
\end{figure*}

The prompts for counselor simulation are shown in Table~\ref{tab:state estimation} to Table~\ref{tab:response rank}. Table~\ref{tab:topic expansion} shows the prompt used to expand the topics and Figure~\ref{fig:topic tree} shows all the collected topics. Table~\ref{tab:counselor state description} to Table~\ref{tab:counselor strategy} demonstrate the descriptions used in counselor agent.

\begin{table*}[tb]
\begin{tabularx}{\textwidth}{X}
\toprule
{\sf \footnotesize During the Motivational Interviewing counseling conversation, the client may exhibit different states that refer to their readiness to change. The client's state can be one of the following: \newline \newline - Precontemplation: The client does not recognize their behavior as problematic and is not considering change. \newline - Contemplation: The client acknowledges the problematic nature of their behavior but is ambivalent about change. \newline - Preparation: The client is ready to take action and is considering steps towards change. \newline \newline Given the current counseling context, analyze the context step by step and then infer the current state of the client. If the context does not clearly indicate the state, it is assumed to be in the Precontemplation state. \newline Your response should be ended with "Therefore, the client's current state in the above context is ..." \newline \newline Given Current Context: \newline [context]  \newline \newline Analyze then Predict State:}
\\ \bottomrule
\end{tabularx}
\caption{Prompt for the counselor agent to infer the state of client. The [context] will be replaced by the conversation so far.}
\label{tab:state estimation}
\end{table*}

\begin{table*}[tb]
\begin{tabularx}{\textwidth}{X}
\toprule
{\sf \footnotesize You are a counselor working with a client to achieve the goal of {self.goal}, specifically addressing the client's behavior, [behavior]. After establishing a foundation of trust, your focus should now shift to identifying specific topics that may motivate the client to change their behavior, [behavior]. \newline \newline In the previous counseling session, the following context was discussed: \newline [context] \newline Your task is to assign the most likely topic(s) from the list below that will engage the client and help them recognize either the benefits of achieving [goal] or the potential risks of continuing with [behavior]. Here are the possible topics you can refer to: \newline [topics] \newline Please respond in JSON format, assigning a probability to each topic in the most top level based on the client's response. For example: {"Health": 0.3, "Economy": 0.2, "Interpersonal Relationships": 0.2, "Law": 0.1, "Education": 0.2} }
\\ \bottomrule
\end{tabularx}
\caption{Prompt for the counselor agent to estimate the probabilities of different topics based on the current context at the beginning of sessions. The module will assign probabilities for each super-class topics based on previous context. The [goal] will be replaced by the counseling goal, such as smoking cessation, reducing alcohol consumption, and the [behavior] will be replaced with the client's problematic behavior. The [context] will be replaced by the conversation so far, while [topics] will be replaced by the super-class topics for counselor agent to choose from.}
\label{tab:topic initialization}
\end{table*}

\begin{table*}[tb]
\begin{tabularx}{\textwidth}{X}
\toprule
{\sf \footnotesize You are acting as a counselor agent, interacting with the client to help them achieve the goal of [goal] related to their behavior, [behavior]. Each client has their own unique motivations, and the counselor help the client discover their inherent motivation for change. The counselor need to explore various topics and uncover what concerns the client the most.  Your task is to analyse the client's feedback toward current topic and then choose the next exploration action (step into, switch, and step out) within the topic tree, but without generating a counselor response. \newline \newline You have already explored the following topics to understand the client’s motivation: \newline [explore\_path] \newline Here is the corresponding counseling conversation: \newline [context] \newline The current topic is [topic]. \newline \newline You have three options: \newline - Step Into: If the client shows interest in this topic, you should dive deeper into its subtopics, including [step\_into\_topics]. \newline - Switch: If the client is interested in the broader category but not this specific topic, switch to another related topic under the same super topic, including[switch\_topics]. \newline - Step Out: If the client’s interest lies in a broader area, step out to a higher-level topic to explore that further, including [step\_out\_topics]. \newline \newline Please analyse the client's feedback toward current situation and then  choose the next course of action based on the feedback without generating counselor's specific response. }
\\ \bottomrule
\end{tabularx}
\caption{Prompt for the counselor agent to explore topics based on the current context. The module will select one action from {\em step into}, {\em switch}, or {\em step out}, then use the corresponding prompt to select the topic. The [goal] will be replaced by the counseling goal, such as smoking cessation, reducing alcohol consumption, and the [behavior] will be replaced with the client's problematic behavior. The [context] will be replaced by the conversation so far, while [explore\_path] and [topic] will be replaced with the explored topics and the current topic. For each action, the placeholders [step\_into\_topics], [switch\_topics], and [step\_out\_topics] will be replaced with the corresponding candidate topics.}
\label{tab:topic exploration}
\end{table*}

\begin{table*}[tb]
\begin{tabularx}{\textwidth}{X}
\toprule
{\sf \footnotesize Your task is to explore the subtopics of the current topic to understand the client's motivation better. \newline \newline You have already explored the following topics to understand the client’s motivation: \newline [explore\_path] \newline Here is the corresponding counseling conversation: \newline [context] \newline The current topic is [topic], and you sense that the client is interested in exploring this further. Please choose one of the following subtopics to dive deeper into the client's motivations: \newline [step\_into\_topics] \newline \newline Analyse the current situation especially for the client's response about the current topic and then select the appropriate next step to motivate the client after analysing. Just analyse for next topic instead of generate response. }
\\ \bottomrule
\end{tabularx}
\caption{Prompt for the counselor agent to step into deeper topics based on the current context. The [context] will be replaced by the conversation so far, while [explore\_path] and [topic] will be replaced by the explored topics and current topic respectively. For each action, the placeholders [step\_into\_topics], [switch\_topics], and [step\_out\_topics] will be replaced by the corresponding candidate topics.}
\label{tab:topic step into}
\end{table*}

\begin{table*}[tb]
\begin{tabularx}{\textwidth}{X}
\toprule
{\sf \footnotesize Your task is to explore a different topic within the same broader category to understand the client's motivation better. \newline \newline You have already explored the following topics to understand the client’s motivation: \newline [explore\_path] \newline Here is the corresponding counseling conversation: \newline [context] \newline and the current client's state is \newline [state]\newline \newline The current topic is [topic], but you sense that the client may be more interested in other topics within the same broader category. Please choose one of the following related topics to continue exploring the client’s motivations: \newline [switch\_topics] \newline \newline Analyse the current situation especially for the client's response about the current topic and then select the appropriate next step to motivate the client after analysing. Just analyse for next topic instead of generate response.} 
\\ \bottomrule
\end{tabularx}
\caption{Prompt for the counselor agent to transit to other topics based on the current context. The [context] will be replaced by the conversation so far and the [state] will be replaced by the inferred client's state. The [explore\_path] and [topic] will be replaced by the explored topics and the current client topic respectively.  For each action, the placeholders [step\_into\_topics], [switch\_topics], and [step\_out\_topics] will be replaced with the corresponding candidate topics.}
\label{tab:topic switch}
\end{table*}

\begin{table*}[tb]
\begin{tabularx}{\textwidth}{X}
\toprule
{\sf \footnotesize Your task is to explore a broader topic to understand the client's motivation better. \newline \newline You have already explored the following topics to understand the client’s motivation: \newline [explore\_path] \newline Here is the corresponding counseling conversation: \newline [context] \newline \newline The current topic is [topic], but you sense that the client may be more interested in another topic instead. Please choose one of the following related topics to continue exploring the client’s motivations: \newline [step\_out\_topics] \newline \newline Analyse the current situation especially for the client's response about the current topic and then select the appropriate next step to motivate the client after analysing. Just analyse for next topic instead of generate response.}
\\ \bottomrule
\end{tabularx}
\caption{Prompt for the counselor agent to step out the current topic and select topic at the parent level based on the current context. The [context] will be replaced by the conversation so far, while [explore\_path] and [topic] will be replaced by the explored topics and current topic respectively. For each action, the placeholders [step\_into\_topics], [switch\_topics], and [step\_out\_topics] will be replaced with the corresponding candidate topics.}
\label{tab:topic step out}
\end{table*}

\begin{table*}[tb]
\begin{tabularx}{\textwidth}{X}
\toprule
{\sf \footnotesize During motivational interviewing, the counselor should employ some counseling strategies tailored to the client's readiness to change, to effectively facilitate behavioral transformation. These counseling strategies are as follows: \newline \newline [strategies] \newline  Based on the current counseling context and the client's state, analyze and select appropriate strategies but no more than 2 for next response to optimally advance the counseling process.  \newline \newline Given Current Context: \newline [context] \newline The current client's state is [state] and you would like to navigate the session toward: [topic] \newline \newline Please analyse the current situation, then select appropriate strategies based on current topic and situation to motivate client after analysing. Remember, you can select up to 2 strategies.}
\\ \bottomrule
\end{tabularx}
\caption{Prompt for the counselor agent to select strategies based on the current context. The [context] will be replaced by the conversation so far while the [state] will be replaced by the inferred state. The [topic] will be replaced by the selected next topic. The [strategies] will be replaced by the list of strategies and their corresponding descriptions.}
\label{tab:strategy selection}
\end{table*}

\begin{table*}[tb]
\begin{tabularx}{\textwidth}{lX}
\toprule
 System Prompt & {\sf \footnotesize You will act as a skilled counselor conducting a Motivational Interviewing (MI) session aimed at achieving [goal] related to the client's behavior, [behavior]. Your task is to help the client discover their inherent motivation to change and identify a tangible plan to change. Start the conversation with the client with some initial rapport building, such as asking, How are you? (e.g., develop mutual trust, friendship, and affinity with the client) before smoothly transitioning to asking about their problematic behavior. Keep the session under 40 turns and each response under 150 characters long. Use the MI principles and techniques described in the Knowledge Base – Motivational Interviewing (MI) context section below. However, these MI principles and techniques are only for you to use to help the user. These principles and techniques, as well as motivational interviewing, should NEVER be mentioned to the user. In each turn, a specific topic will be provided in square brackets after the client's utterance. Guide the conversation toward that topic, ensuring the session explores relevant aspects of the client’s situation and motivation. This will help you tailor your approach to the specific context and steer the counseling toward meaningful insights and actions. \newline \newline Knowledge Base – Motivational Interviewing (MI) \newline Motivational Interviewing (MI) is a counseling approach designed to help individuals find the motivation to make positive behavioral changes. It is widely used in various fields such as health care, addiction treatment, and mental health. Here are the key principles and techniques of Motivational Interviewing: \newline MI Principles \newline - Express Empathy: The foundation of MI is to create a safe and non-judgmental environment where clients feel understood and respected. This involves actively listening and reflecting on what the client is saying, acknowledging their feelings, and showing genuine concern and understanding. \newline ...  \newline MI Techniques \newline At the core of MI are a few basic principles, including expressing empathy and developing discrepancy. Several specific techniques can help individuals make positive life changes from these core principles. Here are some MI techniques that can be used in counseling sessions: \newline - Advise with permission. The counselor gives advice, makes a suggestion, offers a solution or possible action given with prior permission from the client.  \newline - Affirm. The counselor says something positive or complimentary to the client. \newline ...} \\ \midrule
... & ...  \\ \midrule
Assistant & {\sf \footnotesize Counselor: ... }\\ \hline
User & {\sf \footnotesize Client: ... [instruction]}
\\ \bottomrule
\end{tabularx}
\caption{System prompt and generation prompt for the counselor agent. The [behavior] will be replaced by the problematic behavior of client and the [goal] will be replaced by the counseling goal, such as smoking cessation, reducing alcohol consumption. The [instruction] is the turn-level instruction prompt, including the inferred state description, topic description and the given strategy description. In particular, at the beginning of the session, before the counselor has identified a high-probability super-class topic, the instruction will include all super-class topics.}
\label{tab:couneslor system}
\end{table*}

\begin{table*}[tb]
\begin{tabularx}{\textwidth}{X}
\toprule
{\sf \footnotesize You will act as an expert counselor conducting a Motivational Interviewing (MI) session aimed at achieving [goal] related to the client's behavior, [behavior]. Your task is to help the client discover their intrinsic motivation to change and identify a tangible plan for achieving that change. The current state of the counseling session is as follows: \newline [conversation] \newline At this point, multiple responses have been given by peer counselors. Your task is to select the response that best aligns with the current context and adheres to MI principles. Here are the generated responses: \newline [responses] \newline \newline Below are some MI principles for reference: \newline [principles] \newline Please choose the most appropriate response based on the counseling context and the MI principles. Reply with the ID of the response you find most suitable for the current situation.}
\\ \bottomrule
\end{tabularx}
\caption{Prompt for the counselor agent to select the most appropriate response from the candidate responses. The [behavior] placeholder will be replaced with the client's problematic behavior, and the [goal] placeholder will be replaced with the goal of the counseling session. The [responses] placeholder will be replaced with the candidate responses, while the [principles] placeholder will be replaced with the MI principles from psychological literature.}
\label{tab:response rank}
\end{table*}

\begin{table*}[tb]
\begin{tabularx}{\textwidth}{X}
\toprule
{\sf \footnotesize
Your task is to generate additional sub-topics under the category: [category]. Below are some existing sub-topics within this category: \newline [topics] \newline \newline Please suggest new sub-topics under the same category, ensuring they are distinct from the ones provided. Please format your response in the same style.}
\\ \bottomrule
\end{tabularx}
\caption{Prompt for the topic tree expansion. The [category] will be replaced by the parent topic while [topics] will be replaced by the known child topics.}
\label{tab:topic expansion}
\end{table*}

\begin{table*}[tb]
\centering
\begin{tabularx}{\textwidth}{lX}
\toprule
{\bf State}            & {\bf Description }                                           \\ \midrule
Precontemplation & The client is unaware of or underestimates the need for change.       \\ \hline
Contemplation    & The client acknowledges the need for change but remains ambivalent.   \\ \hline
Preparation      & he client is ready to act, planning specific steps toward change.
\\ \bottomrule
\end{tabularx}
\caption{The description of states used in counselor agent.}
\label{tab:counselor state description}
\end{table*}

\begin{table*}[tb]
\centering
\begin{tabularx}{\textwidth}{lX}
\toprule
Topic            & Description                                            \\ \midrule
Infection                   & You can explore how [problematic behavior] increases the risk of infections by weakening the immune system, leading to more frequent or severe infections. You can also discuss how [goal] enhances immune function, reduces infection risks, and improves overall health.                                                                                                                                                         \\ \hline
Hypertension                & You can explore how [problematic behavior] contributes to the development of high blood pressure and increases the risk of complications such as heart disease and stroke. You can also highlight how [goal] helps lower blood pressure, improves heart health, and reduces the risk of cardiovascular conditions.                                                                                                                 \\ \hline
Flu                         & You can explore how [problematic behavior] increases the risk of contracting the flu or experiencing more severe symptoms. You can also explain how [goal] improves immune response, reduces the likelihood of illness, and mitigates the impact of seasonal flu.                                                                                                                                                                  \\ \hline
Inflammation                & You can explore how [problematic behavior] leads to chronic inflammation, increasing the risk of diseases like arthritis, heart disease, or diabetes. You can also highlight how [goal] helps reduce inflammation and supports long-term health.                                                                                                                                                                                   \\ \hline
Liver Disease               & You can explore how [problematic behavior] contributes to liver damage, raising the risk of conditions such as fatty liver disease, cirrhosis, or liver cancer. You can also discuss how [goal] promotes liver health, prevents damage, and reduces the likelihood of chronic liver conditions.                                                                                                                                    \\ \hline
Lung Cancer                 & You can explore how [problematic behavior] increases the risk of lung cancer and other respiratory diseases. You can also emphasize how [goal] lowers the risk of cancer, improves lung function, and enhances overall respiratory health.                                                                                                                                                                                         \\ \hline
COPD                        & You can explore how [problematic behavior] may contribute to the development or worsening of COPD, leading to breathing difficulties and other respiratory issues. You can also discuss how [goal] improves lung function and overall respiratory health.                                                                                                                                                                          \\ \hline
Asthma                      & You can explore how [problematic behavior] triggers or worsens asthma symptoms, increasing the risk of severe attacks. You can also highlight how [goal] helps manage asthma, reduces symptoms, and improves the client's quality of life.                                                                                                                                                                                         \\ \hline
Stroke                      & You can explore how [problematic behavior] increases the risk of stroke, particularly through poor cardiovascular health. You can also discuss how [goal] improves circulation, reduces stroke risk, and supports brain and heart health.                                                                                                                                                                                          \\ \hline
Diabetes                    & You can explore how [problematic behavior] contributes to the development or worsening of diabetes by affecting blood sugar levels. You can also discuss how [goal] helps manage blood sugar, prevent complications, and enhance overall well-being. 
\\ \bottomrule
\end{tabularx}
\caption{The descriptions of topics used in counselor agent (part 1). The [problematic behavior] will be replaced as the client's problematic behavior while the [goal] will be replaced by the counseling goal, such as smoking cessation, reducing alcohol consumption.}
\label{tab:counselor topic description 1}
\end{table*}

\begin{table*}[tb]
\centering
\begin{tabularx}{\textwidth}{lX}
\toprule
Topic            & Description                                            \\ \midrule
Physical Activity           & You can explore how [problematic behavior] reduces physical activity, increasing the risk of obesity, cardiovascular disease, and musculoskeletal issues. You can also emphasize how [goal] increases physical activity and improves overall fitness and health.                                                                                                                                                                   \\ \hline
Sport                       & You can explore how [problematic behavior] reduces performance in sports, limiting physical conditioning and skill development. You can also highlight how [goal] enhances sports participation, improves physical conditioning, and boosts confidence.                                                                                                                                                                            \\ \hline
Physical Fitness            & You can explore how [problematic behavior] negatively affects physical fitness, leading to a decline in overall health. You can also discuss how [goal] promotes better fitness, improves health, and increases energy levels.                                                                                                                                                                                                     \\ \hline
Strength                    & You can explore how [problematic behavior] weakens physical strength, leading to reduced mobility and increased injury risk. You can also discuss how [goal] improves muscle strength, supports healthy aging, and enhances physical performance.                                                                                                                                                                                  \\ \hline
Flexibility                 & You can explore how [problematic behavior] reduces flexibility, increasing stiffness and injury risk. You can also highlight how [goal] improves flexibility, reduces pain, and promotes better movement and posture.                                                                                                                                                                                                              \\ \hline
Endurance                   & You can explore how [problematic behavior] reduces endurance, making it difficult to engage in prolonged physical activities. You can also discuss how [goal] builds endurance, improves stamina, and enhances overall physical performance                                                                                                                                                                                        \\ \hline
Dentistry                   & You can explore how [problematic behavior] leads to poor oral hygiene, increasing the risk of cavities, gum disease, or tooth loss. You can also discuss how [goal] improves oral hygiene, prevents dental problems, and supports overall oral health.                                                                                                                                                                             \\ \hline
Caregiver Burden            & You can explore how [problematic behavior] increases the stress or demands placed on caregivers, leading to burnout and reduced care quality. You can also highlight how [goal] reduces caregiver burden, improves care quality, and supports a healthier caregiving dynamic.                                                                                                                                                      \\ \hline
Independent Living          & You can explore how [problematic behavior] limits a person’s ability to live independently, leading to greater reliance on others for daily needs. You can also emphasize how [goal] promotes independence, improves self-sufficiency, and enhances overall quality of life.                                                                                                                                                       \\ \hline
Human Appearance            & You can explore how [problematic behavior] affects a person’s physical appearance, leading to issues such as skin problems, weight gain, or premature aging. You can also discuss how [goal] improves appearance, boosts self-esteem, and supports overall well-being.                                                                                                                                                             \\ \hline
Depression                  & You can explore how [problematic behavior] worsens symptoms of depression, affecting mood, energy levels, and daily functioning. You can also explain how [goal] improves mental health, enhances mood, and fosters emotional resilience. 
\\ \bottomrule
\end{tabularx}
\caption{The descriptions of topics used in counselor agent (part 2). The [problematic behavior] will be replaced as the client's problematic behavior while the [goal] will be replaced by the counseling goal, such as smoking cessation, reducing alcohol consumption.}
\label{tab:counselor topic description 2}
\end{table*}

\begin{table*}[tb]
\centering
\begin{tabularx}{\textwidth}{lX}
\toprule
Topic            & Description                                            \\ \midrule
Chronodisruption            & You can explore how [problematic behavior] disrupts natural body rhythms, leading to sleep disorders, fatigue, and increased stress. You can also highlight how [goal] restores healthy sleep patterns and improves overall mental and physical health.                                                                                                                                                                            \\ \hline
Anxiety Disorders           & You can explore how [problematic behavior] increases anxiety, leading to chronic stress, panic attacks, or other anxiety-related issues. You can also discuss how [goal] helps manage anxiety, promotes relaxation, and supports emotional well-being.                                                                                                                                                                             \\ \hline
Cognitive Decline           & You can explore how [problematic behavior] accelerates cognitive decline, increasing the risk of dementia and other neurological conditions. You can also highlight how [goal] protects brain health, improves memory, and enhances cognitive function.                                                                                                                                                                            \\ \hline
Safe Sex                    & You can explore how [problematic behavior] increases the risk of sexually transmitted infections (STIs) or unintended pregnancies. You can also explain how [goal] promotes safer sexual practices, reduces health risks, and fosters healthier relationships.                                                                                                                                                                     \\ \hline
Maternal Health             & You can explore how [problematic behavior] impacts maternal health, leading to complications during pregnancy or childbirth. You can also discuss how [goal] supports a healthy pregnancy and reduces the risk of complications.                                                                                                                                                                                                   \\ \hline
Preterm Birth               & You can explore how [problematic behavior] increases the risk of preterm birth, leading to health complications for both mother and baby. You can also highlight how [goal] promotes a healthy pregnancy and reduces the risk of early delivery.                                                                                                                                                                                   \\ \hline
Miscarriage                 & You can explore how [problematic behavior] increases the risk of miscarriage, leading to emotional distress and health complications. You can also emphasize how [goal] supports a healthy pregnancy and reduces the risk of miscarriage.                                                                                                                                                                                          \\ \hline
Birth Defects               & You can explore how [problematic behavior] increases the risk of birth defects. You can also highlight how [goal] supports a healthy pregnancy and reduces the risk of complications.                                                                                                                                                                                                                                              \\ \hline
Productivity                & You can explore how [problematic behavior] negatively affects workplace productivity, leading to decreased performance and career setbacks. You can also explain how [goal] enhances productivity, focus, and career success.                                                                                                                                                                                                      \\ \hline
Absenteeism                 & You can explore how [problematic behavior] strains workplace relationships, leading to conflicts or a negative work environment. You can also discuss how [goal] improves communication, strengthens teamwork, and promotes a positive workplace dynamic. 
\\ \bottomrule
\end{tabularx}
\caption{The descriptions of topics used in counselor agent (part 3). The [problematic behavior] will be replaced as the client's problematic behavior while the [goal] will be replaced by the counseling goal, such as smoking cessation, reducing alcohol consumption.}
\label{tab:counselor topic description 3}
\end{table*}

\begin{table*}[tb]
\centering
\begin{tabularx}{\textwidth}{lX}
\toprule
Topic            & Description                                            \\ \midrule
Workplace Relationships     & Explore how [problematic behavior] may strain workplace relationships, leading to conflicts or a negative work environment. Discuss how achieving [goal] can improve communication, strengthen teamwork, and create a positive workplace dynamic.                                                                                                                                                                                  \\ \hline
Career Break                & You can explore how [problematic behavior] leads to career breaks or job loss, affecting professional growth. You can also highlight how [goal] promotes career continuity and reduces the need for extended leave.                                                                                                                                                                                                                \\ \hline
Career Assessment           & You can explore how [problematic behavior] interferes with career assessments or evaluations, leading to potential setbacks. You can also discuss how [goal] improves career performance and fosters positive evaluations.                                                                                                                                                                                                         \\ \hline
Absence Rate                & You can explore how [problematic behavior] increases the absence rate at work, impacting job security and career progression. You can also highlight how [goal] reduces absences and supports professional growth.                                                                                                                                                                                                                 \\ \hline
Salary                      & You can explore how [problematic behavior] affects salary progression, leading to lower earnings. You can also emphasize how [goal] enhances earning potential and supports financial stability.                                                                                                                                                                                                                                   \\ \hline
Workplace Wellness          & You can explore how [problematic behavior] undermines workplace wellness initiatives, leading to reduced employee health and satisfaction. You can also highlight how [goal] improves workplace wellness and enhances overall job satisfaction.                                                                                                                                                                                    \\ \hline
Workplace Incivility        & You can explore how [problematic behavior] contributes to incivility in the workplace, creating a toxic work environment. You can also discuss how [goal] fosters respect, cooperation, and a positive workplace culture.                                                                                                                                                                                                          \\ \hline
Cost of Living              & You can explore how [problematic behavior] leads to poor financial management, making it harder to meet the cost of living. You can also discuss how [goal] improves financial stability and reduces financial stress.                                                                                                                                                                                                             \\ \hline
Personal Budget             & You can explore how [problematic behavior] makes it difficult to stick to a personal budget, leading to debt and financial challenges. You can also highlight how [goal] improves financial planning and promotes savings.                                                                                                                                                                                                         \\ \hline
Debt                        & You can explore how [problematic behavior] increases debt, impacting credit and financial security. You can also emphasize how [goal] reduces debt and supports financial freedom.                                                                                                                                                                                                                                                 \\ \hline
Income Deficit              & You can explore how [problematic behavior] contributes to income deficits, leading to financial instability. You can also highlight how [goal] improves financial management and reduces income shortfalls. 
\\ \bottomrule
\end{tabularx}
\caption{The descriptions of topics used in counselor agent (part 4). The [problematic behavior] will be replaced as the client's problematic behavior while the [goal] will be replaced by the counseling goal, such as smoking cessation, reducing alcohol consumption.}
\label{tab:counselor topic description 4}
\end{table*}             

\begin{table*}[tb]
\centering
\begin{tabularx}{\textwidth}{lX}
\toprule
Topic            & Description                                            \\ \midrule
Family Estrangement         & You can explore how [problematic behavior] leads to family estrangement, creating emotional distance or separation. You can also explain how [goal] improves family relationships and fosters reconciliation.                                                                                                                                                                                                                      \\ \hline
Family Disruption           & You can explore how [problematic behavior] disrupts family dynamics, leading to conflict and instability. You can also highlight how [goal] strengthens family bonds and promotes harmony.                                                                                                                                                                                                                                         \\ \hline
Divorce                     & You can explore how [problematic behavior] contributes to marital conflict, increasing the risk of divorce. You can also discuss how [goal] improves communication, reduces conflict, and supports a healthy marriage.                                                                                                                                                                                                             \\ \hline
Role Model                  & You can explore how [problematic behavior] negatively influences a parent’s ability to be a positive role model for their children. You can also highlight how [goal] fosters positive behaviors and sets a good example for children.                                                                                                                                                                                             \\ \hline
Child Development           & You can explore how [problematic behavior] affects a child’s development, impacting emotional, social, or cognitive growth. You can also discuss how [goal] supports healthy child development and overall well-being.                                                                                                                                                                                                             \\ \hline
Paternal Bond               & You can explore how [problematic behavior] weakens the paternal bond, leading to strained relationships with children. You can also emphasize how [goal] strengthens the parent-child bond and fosters emotional connection.                                                                                                                                                                                                       \\ \hline
Child Care                  & You can explore how [problematic behavior] interferes with child care, leading to neglect or inconsistency in parenting. You can also highlight how [goal] supports stable, nurturing care and promotes positive outcomes for children.                                                                                                                                                                                            \\ \hline
Habituation                 & You can explore how [problematic behavior] affects a child’s habituation, impacting learning and adaptation. You can also highlight how [goal] promotes healthy habits and learning in children.                                                                                                                                                                                                                                   \\ \hline
Arrest                      & You can explore how [problematic behavior] increases the risk of arrest, leading to legal trouble and a criminal record. You can also highlight how [goal] avoids legal issues and promotes a law-abiding lifestyle.                                                                                                                                                                                                               \\ \hline
Imprisonment                & You can explore how [problematic behavior] increases the risk of imprisonment, with long-term social and legal consequences. You can also explain how [goal] helps avoid incarceration and supports lawful behavior.                                                                                                                                                                                                               \\ \hline
Child Custody               & You can explore how [problematic behavior] impacts a parent’s ability to maintain child custody, leading to legal challenges. You can also highlight how [goal] improves parenting and strengthens legal standing in custody cases.
\\ \bottomrule
\end{tabularx}
\caption{The descriptions of topics used in counselor agent (part 5). The [problematic behavior] will be replaced as the client's problematic behavior while the [goal] will be replaced by the counseling goal, such as smoking cessation, reducing alcohol consumption.}
\label{tab:counselor topic description 5}
\end{table*}   

\begin{table*}[tb]
\centering
\begin{tabularx}{\textwidth}{lX}
\toprule
Topic            & Description                                            \\ \midrule
Traffic Ticket              & You can explore how [problematic behavior] increases the risk of traffic tickets and other legal penalties. You can also discuss how [goal] promotes responsible driving and helps avoid legal infractions.                                                                                                                                                                                                                        \\ \hline
Complaint                   & You can explore how [problematic behavior] increases the likelihood of legal complaints or disputes. You can also discuss how [goal] reduces legal risks and promotes harmonious interactions.                                                                                                                                                                                                                                     \\ \hline
Attendance                  & You can explore how [problematic behavior] affects a student’s attendance, leading to academic challenges and disciplinary actions. You can also highlight how [goal] improves attendance and academic success.                                                                                                                                                                                                                    \\ \hline
Suspension                  & You can explore how [problematic behavior] increases the risk of school suspension, impacting academic progress and relationships. You can also discuss how [goal] reduces suspension risks and supports positive school experiences.                                                                                                                                                                                              \\ \hline
Exam                        & You can explore how [problematic behavior] negatively impacts exam preparation and performance, leading to lower grades. You can also highlight how [goal] improves study habits and exam results.                                                                                                                                                                                                                                 \\ \hline
Scholarship                 & You can explore how [problematic behavior] affects eligibility for scholarships, reducing academic opportunities. You can also emphasize how [goal] improves academic performance and increases scholarship chances.                                                                                                                                                                                                               \\ \hline
Diseases                    & You can explore how [problematic behavior] increases the risk of various diseases, including infections, chronic conditions, and respiratory issues. You can also discuss how [goal] can reduce these risks and support better long-term health. This includes subtopics like infections, hypertension, flu, inflammation, liver disease, lung cancer, chronic obstructive pulmonary disease (COPD), asthma, stroke, and diabetes. \\ \hline
Physical Fitness            & You can explore the negative effects of [problematic behavior] on physical fitness, such as decreased physical activity, loss of strength, and reduced flexibility. You can also discuss how [goal] contributes to better fitness levels, including improvements in endurance, strength, and flexibility.                                                                                                                          \\ \hline
Health Care                 & You can explore how [problematic behavior] affects personal healthcare, such as oral hygiene, independent living, and overall appearance. You can also discuss the positive impact of [goal] in maintaining better health care practices and improving quality of life. Subtopics include dentistry, caregiver burden, independent living, and human appearance.                                                                   \\ \hline
Mental Disorder             & You can explore how [problematic behavior] may contribute to or worsen mental health conditions such as depression, anxiety, and cognitive decline. You can also discuss the benefits of [goal] in managing mental health and improving emotional well-being. Subtopics include depression, chronodisruption, anxiety disorders, and cognitive decline.                                                                            \\ \hline
Sexual Health               & You can explore how [problematic behavior] increases risks related to sexual and reproductive health, such as unsafe sex practices, maternal health complications, and birth defects. You can also discuss how [goal] supports healthier sexual practices and reduces the risk of complications. Subtopics include safe sex, maternal health, preterm birth, miscarriage, and birth defects.                                       
\\ \bottomrule
\end{tabularx}
\caption{The descriptions of topics used in counselor agent (part 6). The [problematic behavior] will be replaced as the client's problematic behavior while the [goal] will be replaced by the counseling goal, such as smoking cessation, reducing alcohol consumption.}
\label{tab:counselor topic description 6}
\end{table*}   

\begin{table*}[tb]
\centering
\begin{tabularx}{\textwidth}{lX}
\toprule
Topic            & Description                                            \\ \midrule
Employment                  & You can explore how [problematic behavior] negatively impacts workplace productivity, absenteeism, and career progress. You can also discuss how [goal] enhances professional success and fosters healthier workplace relationships. Subtopics include productivity, absenteeism, workplace relationships, career break, career assessment, absence rate, salary, workplace wellness, and workplace incivility.                    \\ \hline
Personal Finance            & You can explore how [problematic behavior] leads to financial instability, such as increased debt, poor budgeting, or income deficits. You can also discuss how [goal] helps improve financial management and promotes long-term financial security. Subtopics include cost of living, personal budget, debt, and income deficit.                                                                                                  \\ \hline
Family                      & You can explore how [problematic behavior] leads to family issues, such as estrangement, disruption, or divorce. You can also discuss how [goal] promotes healthier family relationships and reconciliation. Subtopics include family estrangement, family disruption, and divorce.                                                                                                                                                \\ \hline
Parenting                   & You can explore how [problematic behavior] impacts the client's ability to effectively parent, such as being a poor role model or affecting their child’s development. You can also discuss how [goal] enhances positive parenting, strengthens the parent-child bond, and supports healthier child development. Subtopics include role model, child development, paternal bond, child care, and habituation.                      \\ \hline
Criminal Law                & You can explore how [problematic behavior] leads to issues like arrests, imprisonment, or legal complaints. You can also discuss how [goal] helps avoid these legal problems and supports a law-abiding lifestyle. Subtopics include arrest, imprisonment, and complaint.                                                                                                                                                          \\ \hline
Family Law                  & You can explore how [problematic behavior] affects legal matters involving family, such as child custody disputes. You can also discuss how [goal] improves the client’s legal standing and promotes healthier family relationships.                                                                                                                                                                                               \\ \hline
Traffic Law                 & You can explore how [problematic behavior] may lead to traffic violations, such as receiving tickets or facing fines. You can also discuss how [goal] encourages responsible driving and helps avoid legal infractions.                                                                                                                                                                                                            \\ \hline
Student Affairs             & You can explore how [problematic behavior] impacts school attendance, potentially leading to disciplinary actions such as suspension. You can also discuss how [goal] promotes better academic engagement and achievement. Subtopics include attendance, suspension, and scholarship.                                                                                                                                              \\ \hline
Assessment                  & You can explore how [problematic behavior] negatively affects academic performance during assessments, such as exams. You can also discuss how [goal] helps improve focus, study habits, and exam results.                                                                                                                                                                                                                         \\ \hline
Health                      & You can explore how [problematic behavior] impacts your client's physical and mental well-being, leading to potential health issues. You can also discuss the benefits of [goal], which can improve overall quality of life and promote better health outcomes.                                                                                                                                                                    \\ \hline
Economy                     & You can explore how [problematic behavior] affects your client's financial situation, such as through reduced productivity, increased absenteeism, or poor financial management. You can also discuss how [goal] helps improve economic stability and workplace performance.                                                                                                                                                      
\\ \bottomrule
\end{tabularx}
\caption{The descriptions of topics used in counselor agent (part 7). The [problematic behavior] will be replaced as the client's problematic behavior while the [goal] will be replaced by the counseling goal, such as smoking cessation, reducing alcohol consumption.}
\label{tab:counselor topic description 7}
\end{table*}   

\begin{table*}[tb]
\centering
\begin{tabularx}{\textwidth}{lX}
\toprule
Topic            & Description                                            \\ \midrule
Traffic Law                 & You can explore how [problematic behavior] may lead to traffic violations, such as receiving tickets or facing fines. You can also discuss how [goal] encourages responsible driving and helps avoid legal infractions.                                                                                                                                                                                                            \\ \hline
Student Affairs             & You can explore how [problematic behavior] impacts school attendance, potentially leading to disciplinary actions such as suspension. You can also discuss how [goal] promotes better academic engagement and achievement. Subtopics include attendance, suspension, and scholarship.                                                                                                                                              \\ \hline
Assessment                  & You can explore how [problematic behavior] negatively affects academic performance during assessments, such as exams. You can also discuss how [goal] helps improve focus, study habits, and exam results.                                                                                                                                                                                                                         \\ \hline
Health                      & You can explore how [problematic behavior] impacts your client's physical and mental well-being, leading to potential health issues. You can also discuss the benefits of [goal], which can improve overall quality of life and promote better health outcomes.                                                                                                                                                                    \\ \hline
Economy                     & You can explore how [problematic behavior] affects your client's financial situation, such as through reduced productivity, increased absenteeism, or poor financial management. You can also discuss how [goal] helps improve economic stability and workplace performance.        \\ \hline
Interpersonal Relationships & You can explore how [problematic behavior] affects your client’s personal relationships, leading to family strain or issues with parenting. You can also discuss how [goal] strengthens relationships and fosters a healthier family dynamic.                                                                                                                                                                                      \\ \hline
Law                         & You can explore how [problematic behavior] increases legal risks, such as arrests, imprisonment, or traffic violations. You can also discuss how [goal] helps reduce legal troubles and promotes a more responsible approach to law.                                                                                                                                                                                               \\ \hline
Education                   & You can explore how [problematic behavior] interferes with your client’s educational progress, leading to issues like poor attendance, suspension, or missed academic opportunities. You can also discuss how [goal] fosters better academic performance and overall success.
\\ \bottomrule
\end{tabularx}
\caption{The descriptions of topics used in counselor agent (part 8). The [problematic behavior] will be replaced as the client's problematic behavior while the [goal] will be replaced by the counseling goal, such as smoking cessation, reducing alcohol consumption.}
\label{tab:counselor topic description 8}
\end{table*}

\begin{table*}[tb]
\centering
\begin{tabularx}{\textwidth}{lX}
\toprule
Topic            & Description                                            \\ \midrule
Traffic Law                 & You can explore how [problematic behavior] may lead to traffic violations, such as receiving tickets or facing fines. You can also discuss how [goal] encourages responsible driving and helps avoid legal infractions.                                                                                                                                                                                                            \\ \hline
Student Affairs             & You can explore how [problematic behavior] impacts school attendance, potentially leading to disciplinary actions such as suspension. You can also discuss how [goal] promotes better academic engagement and achievement. Subtopics include attendance, suspension, and scholarship.                                                                                                                                              \\ \hline
Assessment                  & You can explore how [problematic behavior] negatively affects academic performance during assessments, such as exams. You can also discuss how [goal] helps improve focus, study habits, and exam results.                                                                                                                                                                                                                         \\ \hline
Health                      & You can explore how [problematic behavior] impacts your client's physical and mental well-being, leading to potential health issues. You can also discuss the benefits of [goal], which can improve overall quality of life and promote better health outcomes.                                                                                                                                                                    \\ \hline
Economy                     & You can explore how [problematic behavior] affects your client's financial situation, such as through reduced productivity, increased absenteeism, or poor financial management. You can also discuss how [goal] helps improve economic stability and workplace performance.        \\ \hline
Interpersonal Relationships & You can explore how [problematic behavior] affects your client’s personal relationships, leading to family strain or issues with parenting. You can also discuss how [goal] strengthens relationships and fosters a healthier family dynamic.                                                                                                                                                                                      \\ \hline
Law                         & You can explore how [problematic behavior] increases legal risks, such as arrests, imprisonment, or traffic violations. You can also discuss how [goal] helps reduce legal troubles and promotes a more responsible approach to law.                                                                                                                                                                                               \\ \hline
Education                   & You can explore how [problematic behavior] interferes with your client’s educational progress, leading to issues like poor attendance, suspension, or missed academic opportunities. You can also discuss how [goal] fosters better academic performance and overall success.
\\ \bottomrule
\end{tabularx}
\caption{The descriptions of topics used in counselor agent (part 9). The [problematic behavior] will be replaced as the client's problematic behavior while the [goal] will be replaced by the counseling goal, such as smoking cessation, reducing alcohol consumption.}
\label{tab:counselor topic description 9}
\end{table*}

\begin{table*}[tb]
\centering
\begin{tabularx}{\textwidth}{lX}
\toprule
Strategy            & Description                                            \\ \midrule
Advise & Give advice, make a suggestion, offer a solution or possible action. For example, "Consider starting with small, manageable changes like taking a short walk daily." \\ \hline
Affirm & Say something positive or complimentary to the client. For example, "You did well by seeking help." \\ \hline
Direct & Give an order, command, direction. The language is imperative. For example, "You’ve got to stop drinking." \\ \hline
Emphasize Control & Directly acknowledges or emphasizes the client's freedom of choice, autonomy, ability to decide, personal responsibility, etc. For example, "It’s up to you to decide whether to drink." \\ \hline
Facilitate & Provide simple utterances that function as "keep going" acknowledgments encouraging the client to keep sharing. For example, "Tell me more about that." \\ \hline
Inform & Give information to the client, explain something, or provide feedback. For example, "This is a hormone that helps your body utilize sugar." \\ \hline
Closed Question & Ask a question in order to gather information, understand, or elicit the client's story. The question implies a short answer: Yes or no, a specific fact, a number, etc. For example, "Did you use heroin this week?" \\ \hline
Open Question & Ask a question in order to gather information, understand, or elicit the client's story. The question should not be closed, and leave latitude for response. For example, "Can you tell me more about your drinking habits?" \\ \hline
Raise Concern & Point out a possible problem with a client's goal, plan, or intention. For example, "What do you think about my plan?" \\ \hline
Confront & Directly disagrees, argues, corrects, shames, blames, seeks to persuade, criticizes, judges, labels, moralizes, ridicules, or questions the client's honesty. For example, "What makes you think that you can get away with it?" \\ \hline
Simple Reflection & Make a statement that reflects back content or meaning previously offered by the client, conveying shallow understanding without additional information. Add nothing at all to what the client has said, but simply repeat or restate it using some or all of the same words. For example, "You don’t want to do that." \\ \hline
Complex Reflection & Make a statement that reflects back content or meaning previously offered by the client, conveying deep understanding with additional information. Change or add to what the client has said in a significant way, to infer the client's meaning. For example, "That’s where you drew the line." \\ \hline
Reframe & Suggest a different meaning for an experience expressed by the client, placing it in a new light. For example, "Maybe this setback is actually a sign that you're ready for change." \\ \hline
Support & Generally supportive, understanding comments that are not codable as Affirm or Reflect. For example, "That must have been difficult for you." \\ \hline
Warn & Provide a warning or threat, implying negative consequences that will follow unless the client takes certain action. For example, "You could go blind if you don’t manage your blood sugar levels." \\ \hline
Structure & Give comments made to explain what is going to happen in the session, to make a transition from one part of a session to another, to help the client anticipate what will happen next. For example, "First, let’s discuss your drinking, and then we can explore other issues." \\ \hline
No Strategy & Say something not related to behavior change. For example, "Good morning!"
\\ \bottomrule
\end{tabularx}
\caption{The descriptions of strategies used in counselor agent. All of them come from~\citet{miller2012motivational}.}
\label{tab:counselor strategy}
\end{table*}

\subsection{Client Agent Implementation}
\label{app:client simulation}
Unlike previous works that predominantly simulate simple client personas~\citep{yosef2024assessing,wu2023towards} or use examples~\citep{chiu2024computational} in LLM prompts, we partially adapt the STAR framework to simulate a client with a few modules that model state transition, dynamic engagement, action selection, and response generation separately. An input {\em client's profile} includes the client's behavioral problem, state of mind, persona, motivation, beliefs, and interests. As the counseling session progresses, the session context (utterance history) is also provided to the client simulator.

During a conversation session, a client is guided through different states to identify changes that can address their behavioral problem. Aligned with the counselor's setting, we also utilize the three states defined in transtheoretical model of health behavior change~\citep{prochaska2005transtheoretical,prochaska2008initial}. The goal of the state transition module is to maintain consistency with how the client may change states during MI counseling, as well as the client's profile. As our work aims to assess the counselor's effectiveness in motivating the client through topic exploration, we focus on the transition between first two states. In the Precontemplation state, the counselor focuses on building trust, establishing change goals, understanding the client's motivations, and addressing misconceptions before evoking a desire to change. The client is expected to enter the Contemplation state only when motivated by specific reason(s) introduced by the counselor. The state transition module thus ensures consistency by analyzing the counselor's utterances to identify mentions of the client's motivations. If such a mention is found, the next state is Contemplation; otherwise, the state remains unchanged.

Each client has specific interests. For example, a parent may be interested in parenting, while a teenager student may be interested in education and friendship. During the engaging stage of MI counseling, the client demonstrates dynamic engagement with topics raised by the counselor, which provides a signal to guide the session. To realistically simulate this behavior, we incorporate a dynamic engagement module. In the client's profile, we designate the fine grained topic ralated to motivation as the ground truth topic. The distance between the current topic and the ground truth topic is used to instruct the client in providing various types of feedback. However, since our topic tree consists of a limited number of topics, the counselor may raise topics that are not covered by the topic tree. To flexibly estimate the distance between topics, we define three topic levels along the path in the topic tree and design four levels of distance, ranging from most distant (not the same super-class topic) to closest (the same fine-grained topic). Based on each level of distance, the client demonstrates different dynamic engagement levels.

Previous works focus on persona but ignore diverse dialogue actions. Moreover, LLMs like ChatGPT are aligned to generate friendly responses~\citep{shen2023large,kopf2024openassistant}. Clients simulated by simple LLM prompting often display overly compliant behavior or a narrow set of actions compared to real human clients~\citep{kang2024can}. To address this challenge, we incorporate an action selection module that uses session context and the client's receptivity (as provided in the profile) to determine appropriate actions. Based on a sample strategy and a receptivity-aware action distribution derived from real data, the client can exhibit realistic behaviors without being overly compliant. Finally, the response generation module takes the state, engagement level, and action as input to generate the response.

The prompt used for client simulation are shown in Table~\ref{tab:state transition}, Table~\ref{tab:dynamic engagement} and Table~\ref{tab:client system}. The Table~\ref{tab:state description}, Table~\ref{tab:action description} and Table~\ref{tab:engagement description} are the descriptions used in client simulation.

\begin{table*}[tb]
\begin{tabularx}{\textwidth}{X}
\toprule
{\sf \footnotesize Your task is to evaluate whether the Counselor's responses align with the Client's motivation concerning a specific topic, target (self or others), and aspect (risk or benefit). Determine if the Counselor's statements effectively motivates the Client. Your analysis should be logical, thorough, and well-supported, providing clear analysis at each step. \newline Here are some examples to help you understand the task better: \newline Here is the conversation snippet toward reducing alcohol consumption: \newline - Counselor: Are you surprised what that might be true? \newline - Client: Yeah, and a couple of my friends drink too. \newline - Counselor: Well, you might not be drinking that much, and other kids are also trying alcohol. I'd like to share with you the risk of using. Alcohol and drugs could really harm you because your brain is still changing. It also-- you're very high risk for becoming addicted. Alcohol and drugs could also interfere with your role in life and your goals, especially in sports, and it could cause unintended sex. How do you feel about this information? \newline The Motivation of Client is as follows: \newline - You are motivated because of the risk of drinking alcohol in sports for yourself, as alcohol would affect your ability to play soccer. \newline Question: Can the Counselor's statement motivate the Client? \newline Analysis: The Counselor's statement addresses various risks associated with alcohol use, including its potential impact on the Client’s role in life and goals, particularly in sports. Since the Client's motivation revolves around the risk of alcohol affecting their ability to play soccer, the Counselor’s mention of how alcohol could interfere with sports aligns with the Client's concern. By highlighting this specific risk, the Counselor's statement effectively taps into the Client’s personal motivation, making it more likely to encourage behavior change. \newline Answer: Yes \newline \newline [other examples]\newline \newline Now, Here is the conversation snippet toward [goal]:\newline - [context] \newline \newline The Motivation of Client is as follows: \newline - [motivation] \newline \newline Question: Can the Counselor's statement motivate the Client?}
\\ \bottomrule
\end{tabularx}
\caption{Prompt for the client simulator to verify the motivation match in a few-shot format. The [other examples] will be replaced by some other real examples annotated by humans. A total of four examples will be used, including two failures and two successes. The [context] will be replaced by the conversation so far, and the [motivation] will be replaced by the motivation of client. The [goal] will be replaced by the counseling goal, such as smoking cessation, reducing alcohol consumption.}
\label{tab:state transition}
\end{table*}

\begin{table*}[tb]
\begin{tabularx}{\textwidth}{X}
\toprule
{\sf \footnotesize You are provided with a dialogue context from a counseling session and a specific target topic. Your task is to evaluate whether the counselor's statements relate to the given topic. Analyze the session and the counselor's responses to determine if they proactively mention the target topic explicitly.
- If the counselor’s statements mention the provided topic explicitly, respond with "Yes." \newline - If the counselor’s statements do not mention the provided topic explicitly, respond with "No." \newline \newline Your analysis should focus on whether the counselor has appropriately captured and addressed the target topic.\newline [examples]\newline \newline \newline Now, Here is the conversation snippet:\newline - [context] \newline \newline The Concerned Topic is as follows: \newline - [topic] \newline \newline Question: Do the counselor’s focus match the topics of concern given?}
\\ \bottomrule
\end{tabularx}
\caption{Prompt for the client simulator to compute the distance between the current topic in the given context and the concerned topic in a few-shot format. The [examples] will be replaced by real examples annotated by humans, the [context] will be replaced by the conversation so far, and the [topic] will be replaced by the client's topic of interest. Since there are multiple levels of topics, the distance will be computed from the most fine-grained topic up, allowing the module to match the most specific topic. The computed distance will be used to derived the engagement level.}
\label{tab:dynamic engagement}
\end{table*}

\begin{table*}[tb]
\begin{tabularx}{\textwidth}{X}
\toprule
{\sf \footnotesize Assume you are a Client involved in a counseling conversation. The current conversation is provided below: \newline [context] \newline \newline Based on the context, allocate probabilities to each of the following dialogue actions to maintain coherence:  \newline [actions] \newline Provide your response in JSON format, ensuring that the sum of all probabilities equals 100. For example: {'Deny': 35, 'Downplay': 25, 'Blame': 25, 'Inform': 5, 'Engage': 10}}
\\ \bottomrule
\end{tabularx}
\caption{Prompt for the client simulator to provide the probability to select the action. The [context] will be replaced by previous context while [actions] will be replaced by actions set in corresponding to current state.}
\label{tab:action selection}
\end{table*}

\begin{table*}[tb]
\begin{tabularx}{\textwidth}{lX}
\toprule
System Prompt & {\sf \footnotesize In this role-play scenario, you'll take on the role of a Client discussing about your [behavior]. \newline Here is your personas which you need to follow consistently throughout the conversation:\newline [personas] \newline Here is a conversation occurs in parallel world between you (Client) and Counselor, where you can follow the style and information provided in the conversation: \newline [reference] \newline \newline Please follow these guidelines in your responses: \newline - Start your response with "Client: " \newline - Adhere strictly to the state, action and persona specified within square brackets. \newline - Keep your responses coherent and concise, similar to the reference conversation and no more than 3 sentences. \newline - Be natural and concise without being overly polite. \newline - Stick to the persona provided and avoid introducing contradictive details} \\ \midrule
... & ... \\ \midrule
Assistant & {\sf \footnotesize Client:} ... \\ \hline
User & {\sf \footnotesize Counselor: ... [instruction]}
\\ \bottomrule
\end{tabularx}
\caption{System prompt and generation prompt for the client simulator. The placeholder {\sf \footnotesize [behavior]} will be replaced by the problematic behavior of client. {\sf \footnotesize [Personas]} will be replaced by the personas of client, while {\sf \footnotesize [reference]} will be replaced by the original counseling session. {\sf \footnotesize  [instruction]} is the turn-level instruction prompt, including the state description, action description and topic matching outcome description (from Table 35).}
\label{tab:client system}
\end{table*}

\begin{table*}[tb]
\centering
\begin{tabularx}{\textwidth}{lXp{4cm}}
\toprule
State            & Description                                                                                                                                  & Corresponding Actions                                    \\ \midrule
Precontemplation & The client is unaware of or underestimates the need for change.           & Inform, Engage, Deny, \newline Blame, Downplay     \\ \hline
Contemplation    & The client acknowledges the need for change but remains ambivalent.     & Inform, Engage, Hesitate, \newline  Doubt, Acknowledge   
\\ \bottomrule
\end{tabularx}
\caption{The states of change and corresponding actions used in client simulation.}
\label{tab:state description}
\end{table*}

\begin{table*}[tb]
\centering
\begin{tabularx}{\textwidth}{lX}
\toprule
Action      & Description                                                                                                                                      \\ \midrule
Deny        & The client directly refuses to admit their behavior is problematic or needs change.                                                              \\ \hline
Downplay    & The client downplays the importance or impact of their behavior or situation.                                                                    \\ \hline
Blame       & The client attributes their issues to external factors, such as stressful life or other people.                                                  \\ \hline
Hesitate    & The client shows uncertainty, indicating ambivalence about change.                                                                               \\ \hline
Doubt       & The client expresses skepticism about the practicality or success of proposed changes.                                                           \\ \hline
Engage      & The client interacts politely with the counselor, such as greeting, thanking or ask questions.                                                   \\ \hline
Inform      & The client shares details about their background, experiences, or emotions.                                                                      \\ \hline
Acknowledge & The client highlight the importance, benefit or confidence to change.                                                                            \\ \hline
Accept      & The client agrees to adopt the suggested action plan.                                                                                            \\ \hline
Reject      & The client declines the proposed plan, deeming it unsuitable.                                                                                    \\  \hline
Plan        & The client proposes or details steps for a change plan.                                                                                          \\ \hline
Terminate   & The client highlights current state, expresses a desire to end the current session, and suggests further discussion be deferred to a later time. \\ \bottomrule
\end{tabularx}
\caption{Descriptions of each action used in client simulation.}
\label{tab:action description}
\end{table*}

\begin{table*}[tb]
\centering
\begin{tabularx}{\textwidth}{lX}
\toprule
Topic Matching Outcome          & Description                                                                                                                                      \\ \midrule
Different SuperClass Topics        & You should provide vague and broad answers that avoid focusing on the current topic. Shift the conversation subtly toward unrelated areas, without engaging deeply with the topic.                      \\ \hline
Same SuperClass, Different Coarse-Grained Topics    & Acknowledge the general relevance of the topic, but hint that your focus lies elsewhere within the broad category.                                       \\ \hline
Same Coarse-Grained, Different Fine-Grained Topics       & Engage more directly with the topic. Offer responses that hint there’s a deeper, more specific issue to explore.                                                  \\ \hline
Same Fine-Grained Topic    & Provide specific responses that affirm the counselor is on the right track. Offer deeper insights and confirm the relevance of the topic, fully engaging with the conversation.                \\ \bottomrule
\end{tabularx}
\caption{Matching outcomes between current session context and client's interested motivation topic and the descriptions of the matching outcomes used in client simulation (see Table 32).}
\label{tab:engagement description}
\end{table*}

\subsection{Moderator Implementation}

As shown in Table~\ref{tab:moderator prompt}, the moderator determines the conclusion of the counseling sessions based on three conditions using a few-shot prompt for GPT-4o.

\begin{table*}[tb]
\begin{tabularx}{\textwidth}{X}
\toprule
{\sf \footnotesize
Your task is to assess the current state of the conversation (the most recent utterances) and determine whether the conversation has concluded.\newline The conversation is considered to have concluded if any of the following conditions are met:\newline
- The Client or Counselor explicitly expresses to end the conversation \newline
- The Counselor successfully motivates the Client and the Client proactively acknowledges to change.\newline
- The Counselor decides not to pursue any changes in the Client's behavior and communicates readiness to provide support in the future.
\newline \newline Here are some examples to help you understand the task better:\newline [examples]\newline \newline \newline Here is a new Conversation Snippet:\newline [context]\newline \newline Question: Should the conversation be concluded?}  \\ \bottomrule
\end{tabularx}
\caption{Prompt for the moderator in a few-shot Format. The [examples] will be replaced by real examples annotated by human and the [context] will be replaced by the conversation so far.}
\label{tab:moderator prompt}
\end{table*}

\subsection{MITI Annotation Implementation}
\label{app:miti annotation}
Although LLMs demonstrate superior performance in language understanding, they do not perform well in annotating labels of strategies used in counselor's utterances. The accuracy results of GPT-4o in both AnnoMI~\citep{wu2022anno} and another open-source dataset~\citep{welivita2022curating} are below 60\% (53.6\% and 47.2\%, respectively). We instead follow prior work on annotating MI dialogues~\citep{shah2022modeling} and using them to fine-tune a BERT~\citep{kenton2019bert} classifier which predicts the MI behavior codes for each counselor utterance. We train the classifier on a publicly available dataset~\citep{welivita2022curating} with annotated strategies specific to MI.  We do not use AnnoMI~\citep{wu2022anno} because it does not contain fine-grained and other topic labels. We use a 0.8-0.1-0.1 train-validation-test split. We fine-tune BERT using the Huggingface Transformers Library~\citep{wolf2020transformers} for a maximum of 10 epochs, with early stopping option and a learning rate of 5e-5.

However, the accuracy of the fine-tuned BERT classifier is still not optimal (only 56.2\% accuracy). Nevertheless, we found that the recall@5 across all classes was high (89.34\%). Therefore, we employed the fine-tuned BERT classifier to retrieve the top 5 possible behavior labels, followed by GPT-4o to determine the final label. This approach yielded an acceptable accuracy of 72.8\%. The prompt used for GPT-4o is shown in Table~\ref{tab:miti competence prompt}

\begin{table*}[tb]
\begin{tabularx}{\textwidth}{X}
\toprule
{\sf \footnotesize
You are a motivational interviewing assistant tasked with MITI Behavioral Code Annotation for the counselor's utterance based on previous context and the provided utterance. \newline \newline Here are the MITI Behavioral Codes for your reference: \newline - Advise with Permission: Offer advice, suggestions, or possible actions after obtaining the client’s permission. For example, "Consider starting with small, manageable changes like taking a short walk daily." \newline  - Advise without Permission: Give advice, makes a suggestion, offers a solution or possible action without the permission of client. For example, "You could ask your friends not to bring drugs when they come over." \newline ... (this part is removed for space conservation) ... \newline - Filter: Say something not related to behavior change. For example, "Good morning!" \newline  \newline  \newline Context from the previous session: \newline [context] \newline  \newline Counselor's utterance to annotate: \newline [utterance] \newline Please select the most appropriate MITI Behavioral Code for the given counselor’s utterance. While multiple codes may apply, choose only the primary one.
}\\
\bottomrule
\end{tabularx}
\caption{Prompt for GPT-4o to annotate the MITI behavior code for a given utterance. The [context] placeholder will be replaced with the previous session context, and the [utterance] placeholder will be replaced with the utterance to be annotated.}
\label{tab:miti competence prompt}
\end{table*}

\section{Client Experience Evaluation}
\label{sec:client evaluation}

In addition to MI competence, client experience is also widely used to evaluate counseling sessions. Following previous work~\citep{wang2024towards}, we evaluate session outcomes, therapeutic alliance, and self-reported feelings using questionnaires such as the Client Evaluation of Counselor Scale~\citep{hamilton2000construct} and the Working Alliance Inventory - Short Revised (WAI-SR;\citep{hatcher2006development}). We employ GPT-4o with the prompt from\citet{wang2024towards} to provide scores for each generated session.

\begin{table}[tb]
\resizebox{0.49\textwidth}{!}{
\begin{tabular}{lrrrrrr}
\toprule
     & SO$\uparrow$ & TA$\uparrow$          & Depth$\uparrow$         & Smoothness$\uparrow$    & Positivity$\uparrow$    & Arousal$\uparrow$       \\ \midrule
HQ & 0.76            & 0.78               & 4.21          & 5.13          & 4.67          & 4.01          \\
LQ  & 0.41            & 0.48               & 3.43          & 3.11          & 3.60          & 3.63          \\ \midrule
\multicolumn{7}{c}{\cellcolor[HTML]{E0E0E0}GPT-4o Based Counselor}                                                 \\
Base & 0.66            & 0.69               & 3.26          & 4.33          & 4.30          & 3.47          \\
DIIR & 0.65            & 0.68               & 3.32          & 4.17          & 4.13          & 3.48          \\
CoS  & 0.68            & 0.71               & 3.37          & 4.42          & 4.31          & 3.60          \\
CAMI-TE  & 0.69         & 0.71               & 3.42          & 4.47          & 4.35          & 3.63          \\
CAMI & \textbf{0.72}   & \textbf{0.75}      & \textbf{3.63} & \textbf{4.82} & \textbf{4.43} & \textbf{3.71} \\ \midrule
\multicolumn{7}{c}{\cellcolor[HTML]{E0E0E0}Llama-3.1 70B Based Counselor}                                          \\
Base & 0.58            & 0.61               & 2.93          & 3.24          & 3.71          & 3.12          \\
CoS  & 0.63            & 0.63               & 3.08          & 3.40          & 3.84          & 3.23          \\
DIIR & 0.61            & 0.57               & 3.02          & 3.08          & 3.37          & 3.18          \\
CAMI-TE  & 0.61         & 0.59               & 3.09          & 3.13          & 3.49          & 3.23          \\
CAMI & \textbf{0.68}   & \textbf{0.68}      & \textbf{3.47} & \textbf{3.88} & \textbf{3.97} & \textbf{3.66}          \\ \bottomrule
\end{tabular}}
\caption{Results of the client experience assessment, including session outcome (SO), therapeutic alliance (TA), and self-reported feelings scores. The self-reported feelings scores include Depth, Smoothness, Positivity, and Arousal. }
\label{tab:client assessment}
\end{table}

As shown in Table~\ref{tab:client assessment}, CAMI achieves higher client assessment compared to other baselines. However, compared to the high-quality human counseling sessions, there is still significant room for improvement. Furthermore, the performance of some baselines falls below that of low-quality sessions. This indicates that while LLM-based counselor agents may demonstrate MI technique competence, there are other aspects that need to be addressed to further improve the client experience.

\section{Experts Evaluation}
\label{app:expert evaluation}

We instructed professional experts to annotate the given conversations from multiple aspects, including MITI rating, Change Talk Exploration, Success in Eliciting Change Talk, Counselor Realism, and Client Realism. The MITI rating assesses the counselor's behavior during the observed session, including Cultivating Change Talk (Table~\ref{tab:cultivating change talk score}), Softening Sustain Talk (Table~\ref{tab:softening sustain talk score}), Partnership (Table~\ref{tab:partnership score}), and Empathy (Table~\ref{tab:empathy score}), with each item scored on a 1-5 scale. Change Talk Exploration evaluates the counselor's capability to explore motivation topics in the right direction, with items rated on a 5-point scale (Table~\ref{tab:change talk exploration score}). Furthermore, Success in Eliciting Change Talk assesses whether the counselor successfully motivates the client, similar to the automatic evaluation of success rate, and is rated on a three-point scale (Table~\ref{tab:success in eliciting change talk score}). Finally, we instructed experts to evaluate the realism of the counselor (Table~\ref{tab:counselor realism}) and client (Table~\ref{tab:client realism}) based on language, tone, and responses, to assess the effectiveness of the counselor and the consistency of the client. These two items are rated on a five-point scale.

\begin{table*}[tb]
\begin{tabularx}{\textwidth}{X}
\toprule
\textbf{Cultivating Change Talk} \\ \midrule
1: Clinician shows no explicit attention to, or preference for, the client’s language in favor of changing.                                                                                               \\
2: Clinician sporadically attends to client language in favor of change – frequently misses opportunities to encourage change talk.                                                                       \\
3: Clinician often attends to the client’s language in favor of change, but misses some opportunities to encourage change talk.                                                                           \\
4: Clinician consistently attends to the client’s language about change and makes efforts to encourage it.                                                                                                \\
5: Clinician shows a marked and consistent effort to increase the depth, strength, or momentum of the client’s language in favor of change.                                                               \\
\bottomrule                            
\end{tabularx}
\caption{Cultivating Change Talk Scores and Descriptions.}
\label{tab:cultivating change talk score}
\end{table*}

\begin{table*}[tb]
\begin{tabularx}{\textwidth}{X}
\toprule
\textbf{Softening Sustain Talk} \\ \midrule
1: Clinician consistently responds to the client’s language in a manner that facilitates the frequency or depth of arguments in favor of the status quo.                                            \\
2: Clinician usually chooses to explore, focus on, or respond to the client’s language in favor of the status quo.        \\
3: Clinician gives preference to the client’s language in favor of the status quo, but may show some instances of shifting the focus away from sustain talk.                                           \\
4: Clinician typically avoids an emphasis on client language favoring the status quo.                                            \\
5: Clinician shows a marked and consistent effort to decrease the depth, strength, or momentum of the clients language in favor of the status quo.                                            \\
\bottomrule                            
\end{tabularx}
\caption{Softening Sustain Talk Scores and Descriptions.}
\label{tab:softening sustain talk score}
\end{table*}

\begin{table*}[tb]
\begin{tabularx}{\textwidth}{X}
\toprule
\textbf{Partnership} \\ \midrule
1: Clinician actively assumes the expert role for the majority of the interaction with the client. Collaboration or partnership is absent.                                      \\
2: Clinician superficially responds to opportunities to collaborate.        \\
3: Clinician incorporates client’s contributions but does so in a lukewarm or erratic fashion.                                         \\
4: Clinician fosters collaboration and power sharing so that client’s contributions impact the session in ways that they otherwise would not. \\
5: Clinician actively fosters and encourages power sharing in the interaction in such a way that client’s contributions substantially influence the nature of the session.                                \\
\bottomrule                            
\end{tabularx}
\caption{Partnership Scores and Descriptions.}
\label{tab:partnership score}
\end{table*}

\begin{table*}[tb]
\begin{tabularx}{\textwidth}{X}
\toprule
\textbf{Empathy} \\ \midrule
1: Clinician gives little or no attention to the client’s perspective.                  \\
2: Clinician makes sporadic efforts to explore the client’s perspective. Clinician’s understanding may be inaccurate or may detract from the client’s true meaning.      \\
3: Clinician is actively trying to understand the client’s perspective, with modest success. \\
4: Clinician makes active and repeated efforts to understand the client’s point of view. Shows evidence of accurate understanding of the client’s worldview, although mostly limited to explicit content. \\
5: Clinician shows evidence of deep understanding of client’s point of view, not just for what has been explicitly stated but what the client means but has not yet said. \\
\bottomrule                            
\end{tabularx}
\caption{Empathy Scores and Descriptions}
\label{tab:empathy score}
\end{table*}

\begin{table*}[tb]
\begin{tabularx}{\textwidth}{X}
\toprule
\textbf{Motivation Topic Exploration} \\ \midrule
1: Counsellor fails to explore.                  \\
2: Counsellor tried but was not effective in determining the right motivation topic.      \\
3: Counsellor tried but was partially effective. \\
4: Counsellor is close to determining the right motivation topic. \\
5: Counsellor successfully determines the right motivation topic. \\
\bottomrule                            
\end{tabularx}
\caption{Motivation Topic Exploration Scores and Descriptions}
\label{tab:change talk exploration score}
\end{table*}

\begin{table*}[tb]
\begin{tabularx}{\textwidth}{X}
\toprule
\textbf{Success in Eliciting Change Talk} \\ \midrule
1: Failure in eliciting change talk.                  \\
2: Partial success in eliciting change talk.      \\
3: Success in eliciting change talk. \\
\bottomrule                            
\end{tabularx}
\caption{Success in Eliciting Change Talk Scores and Descriptions}
\label{tab:success in eliciting change talk score}
\end{table*}

\begin{table*}[tb]
\begin{tabularx}{\textwidth}{X}
\toprule
\textbf{Counselor Realism} \\ \midrule
1 (Highly Unrealistic): Language, tone, and responses are completely mechanical, lacking empathy or relevance. The counselor's responses are not adapted to client input at all.                  \\
2 (Somewhat Unrealistic): Language, tone, and responses are often robotic, repetitive, or overly generalized, with limited adaptation to client input.      \\
3 (Moderately Realistic): Language, tone, and responses are mostly accurate and somewhat conversational but often mechanical. The counselor may miss emotional cues and occasionally lapse into generic advice or inconsistent empathy. \\
4 (Mostly Realistic): Language, tone, and responses are reflective of a human counselor with occasional minor inconsistencies, mechanical phrasing, or lack of emotional nuance. \\
5 (Highly Realistic): Language, tone, and responses are indistinguishable from a human counselor. The counselor's responses are empathetic and personalized to client input. \\
\bottomrule                            
\end{tabularx}
\caption{Counselor Realism Scores and Descriptions}
\label{tab:counselor realism}
\end{table*}

\begin{table*}[tb]
\begin{tabularx}{\textwidth}{X}
\toprule
\textbf{Client Realism} \\ \midrule
1 (Highly Unrealistic): Language, tone, and responses are completely mechanical, lacking any emotional depth or relevance to the client's background and stage of change. The client's responses do not resemble those of a real person, showing no awareness of context or emotional engagement.                  \\
2 (Somewhat Unrealistic): Language, tone, and responses are often robotic or repetitive, showing limited emotional nuance. Attempts to align the client's responses with the background and state of change are poorly executed.      \\
3 (Moderately Realistic): Language, tone, and responses mostly align with the client's background and stage of change but often lack variability or emotional depth. The client's responses may feel too predictable or exhibit excessive compliance or resistance. \\
4 (Mostly Realistic): Language, tone, and responses are believable, with occasional minor inconsistencies, unnatural phrasing, or a lack of emotional depth in relation to the client's background and stage of change. \\
5 (Highly Realistic): Language, tone, and responses are indistinguishable from a human client. The client's responses are complex and express emotions that are appropriate to the client's background and stage of change. \\
\bottomrule                            
\end{tabularx}
\caption{Client Realism Scores and Descriptions}
\label{tab:client realism}
\end{table*}

\section{Example}
\label{app:example}

\begin{figure*}
    \centering
    \includegraphics[width=\linewidth]{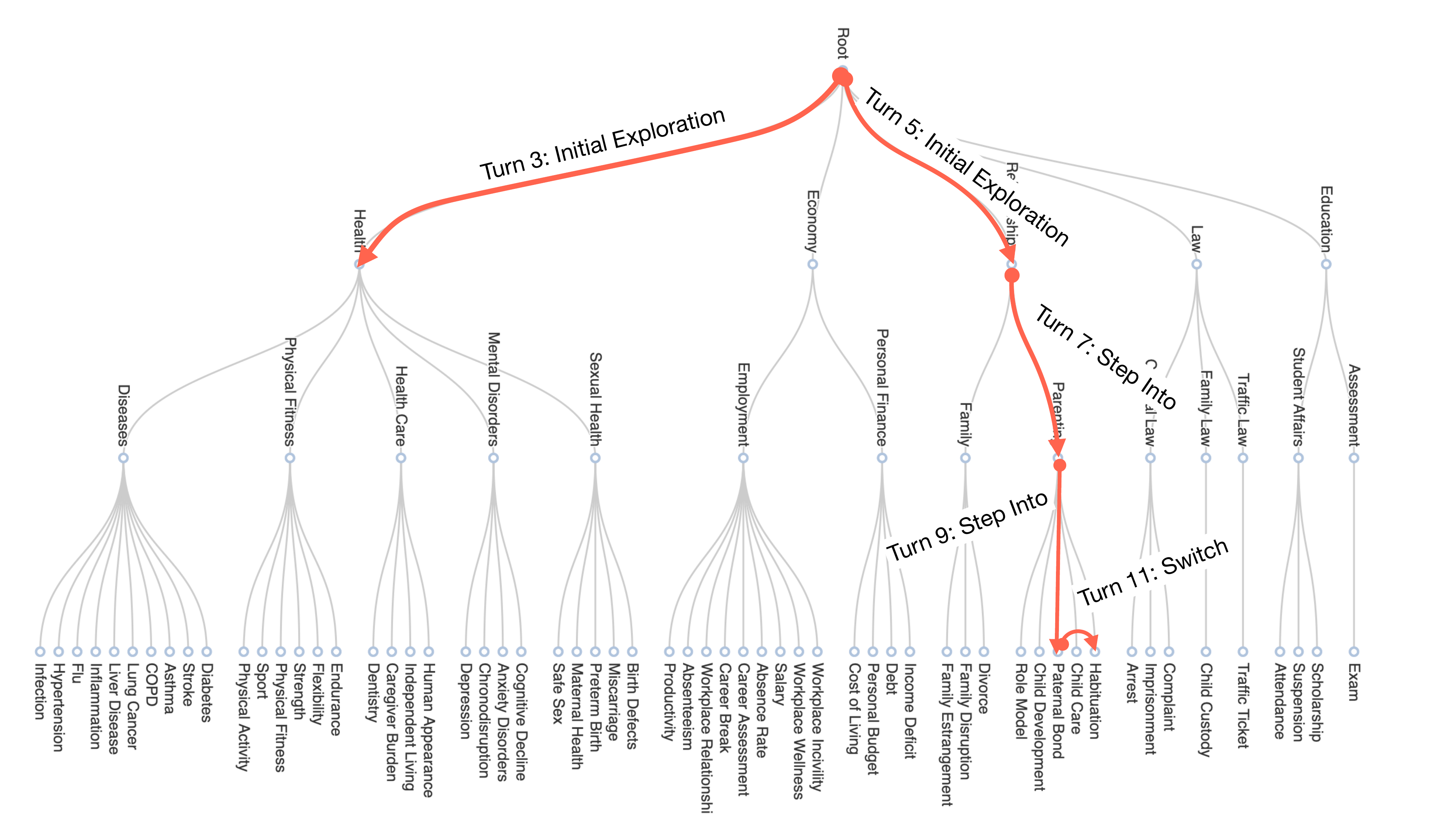}
    \caption{Topic Exploration Path by the Counselor in the Example.}
    \label{fig:exploration path}
\end{figure*}

Table~\ref{tab:example} shows a counseling session between a client agent and CAMI, with the client's profile shown in Table~\ref{tab:example profile}. During the session, CAMI explores Health and Relationships topics in the first 6 turns.  It subsequently performs Step Into operation from Relationships to Parenting topics in Turn 7, and from Parenting to Paternal Bond topics in Turn 9, before a Switch operation to Habituation topic in Turn 11.  The full exploration path is visualized in Figure~\ref{fig:exploration path}.

\begin{table*}[ht]
\centering
\begin{tabularx}{\textwidth}{c X}
\toprule
        {\bf Turn ID} & {\bf Utterances}                     \\ \midrule
1       & Counselor: Hello. How are you?                                                                                                                                                                                                                                                                                                                     \\ \hline
2       & Client: I am good. What about you?                                                                                                                                                                                                                                                                                                                 \\ \hline
3       & Counselor: I'm doing well, thanks for asking! Can you share how you've been feeling about your health lately, especially in relation to smoking?                                                                                                                                                                                                   \\ \hline
4       & Client: I've been pretty busy with work and juggling family stuff. Sometimes I just think about how nice it is to take a break with my coworkers outside, chatting and during those moments, life feels a bit more relaxed. You know how it is—it’s nice to unwind.                                                                                \\ \hline
5       & Counselor: It sounds like smoking is a comforting routine for you amidst your busy life with work and your daughter. I wonder how it might impact the time you spend with your family.                                                                                                                                                             \\ \hline
6       & Client: I get what you mean, but I really don't see my smoking as an issue. It helps me manage stress, and I don’t think it affects my time with my daughter. We still have good moments together, and I wouldn't want to change that.                                                                                                             \\ \hline
7       & Counselor: It sounds like you find comfort in smoking to manage stress, and that feels crucial to your parenting now. I wonder, though, how you might feel if your daughter started to see smoking as a way to unwind, too. What impact would that have on your relationship with her?                                                             \\ \hline
8       & Client: That’s a tough thought. I remember starting young and how hard it was to quit later. I wouldn’t want that for her. It makes me reflect on what I’m modeling.                                                                                                   \\ \hline
9       & Counselor: It's great that you're reflecting on your modeling for your daughter. Perhaps seeing smoking as something that may weaken your bond could help reshape how you approach stress—considering healthier ways to connect can actually strengthen your relationship and set a positive example for her.                                      \\ \hline
10      & Client: I appreciate that perspective. I do want to set a strong example for her. My experience with my own family and their struggles makes me cautious. I see the patterns and don’t want her to feel that smoking is a go-to solution for stress, like I did.                                                                                   \\ \hline
11      & Counselor: It sounds like your family’s struggles have really shaped your perspective on smoking. You want to protect your daughter from those same patterns, knowing that smoking can not only impact her stress management but also her overall learning and adaptation to challenges. That’s a powerful motivation.                             \\ \hline
12      & Client: Exactly. I want her to learn healthier coping strategies instead of following in my footsteps. I realize that I need to make a change for both our sakes. It’s not just about me; it’s about creating an environment where she can thrive without those habits. It's time to find better ways to manage stress that we can share together.   
\\ \bottomrule
\end{tabularx}
\caption{Counseling Session between CAMI and a Simulated Client.}
\label{tab:example}
\end{table*}

\begin{table*}[ht]
\centering
\begin{tabularx}{\textwidth}{l X}
\toprule
{\bf Profile Item}        & {\bf Description}                     \\ \midrule
Behavioral Problem  & Smoking \\ \hline
State of Mind       & Precontemplation \\ \hline
Personas            & You smoked when you were young.", "You have a 12-year-old daughter. \newline The friends of the your daughter have tried smoking. \newline When you are at work, you all go out and smoke and chit chat and talk about things. \newline You like to have a cigarette after eating. \\ \hline
Motivation          & You are motivated because of the risk of smoking in habituation for your daughter, as you don't want your daughter to think smoking is good thing.    \\ \hline
Beliefs             & You feel smoking enjoyable and relaxing. \newline You would be left out if you would hate to not smoke. \newline You use smoking as a way to manage weight and control eating habits. \newline You have been smoking for a long time and feels it's almost impossible to quit. \\ \hline
Motivation Topics          & Interpersonal Relationships, Parenting, Habituation
\\ \bottomrule
\end{tabularx}
\caption{Client's Profile in the Example.}
\label{tab:example profile}
\end{table*}

\end{document}